\documentclass[10pt,twocolumn,letterpaper]{article}

\usepackage[pagenumbers]{cvpr} 

\usepackage{graphicx}
\usepackage{amsmath}
\usepackage{amssymb}
\usepackage{booktabs}
\usepackage{multirow}
\usepackage{arydshln}
\usepackage{times}
\usepackage{epsfig}
\usepackage{multicol}
\usepackage{verbatim}

\usepackage[pagebackref,breaklinks,colorlinks]{hyperref}

\usepackage[capitalize]{cleveref}
\crefname{section}{Sec.}{Secs.}
\Crefname{section}{Section}{Sections}
\Crefname{table}{Table}{Tables}
\crefname{table}{Tab.}{Tabs.}

\begin{document}

\title{Where We Are and What We're Looking At: Query Based Worldwide Image Geo-localization Using Hierarchies and Scenes}

\author{Brandon Clark\thanks{These authors contributed equally to the work}, Alec Kerrigan$^*$, Parth Parag Kulkarni, Vicente Vivanco Cepeda, Mubarak Shah\\
Center for Research in Computer Vision, University of Central Florida, Orlando, USA\\
{\tt\small \{brandonclark314, aleckerrigan, parthpkulkarni.pk, vicente.vivanco\}@knights.ucf.edu,}\\
\tt\small shah@crcv.ucf.edu}
\maketitle
\begin{abstract}
   \vspace{-0.6cm}
   Determining the exact latitude and longitude that a photo was taken is a useful and widely applicable task, yet it remains exceptionally difficult despite the accelerated progress of other computer vision tasks. 
   Most previous approaches have opted to learn a single representation of query images, which are then classified at different levels of geographic granularity. These approaches fail to exploit the different visual cues that give context to different hierarchies, such as the country, state, and city level. To this end, we introduce an end-to-end transformer-based architecture that exploits the relationship between different geographic levels (which we refer to as hierarchies) and the corresponding visual scene information in an image through \it{hierarchical cross-attention}. We achieve this by learning a query for each geographic hierarchy and scene type. Furthermore, we learn a separate representation for different environmental scenes, as different scenes in the same location are often defined by completely different visual features. We achieve state of the art street level accuracy on 4 standard geo-localization datasets : Im2GPS, Im2GPS3k, YFCC4k, and YFCC26k, as well as qualitatively demonstrate how our method learns different representations for different visual hierarchies and scenes, 
   which has not been demonstrated in the previous methods. These previous testing datasets mostly consist of iconic landmarks or images taken from social media, which makes them either a memorization task, or biased towards certain places. To address this issue we introduce a much harder testing dataset, \it{Google-World-Streets-15k}, comprised of images taken from Google Streetview covering the whole planet and present state of the art results. Our code will be made available in the camera-ready version. 
   
   
\end{abstract}


\begin{figure*}[h]
    \centering
    \includegraphics[width=1 \textwidth]{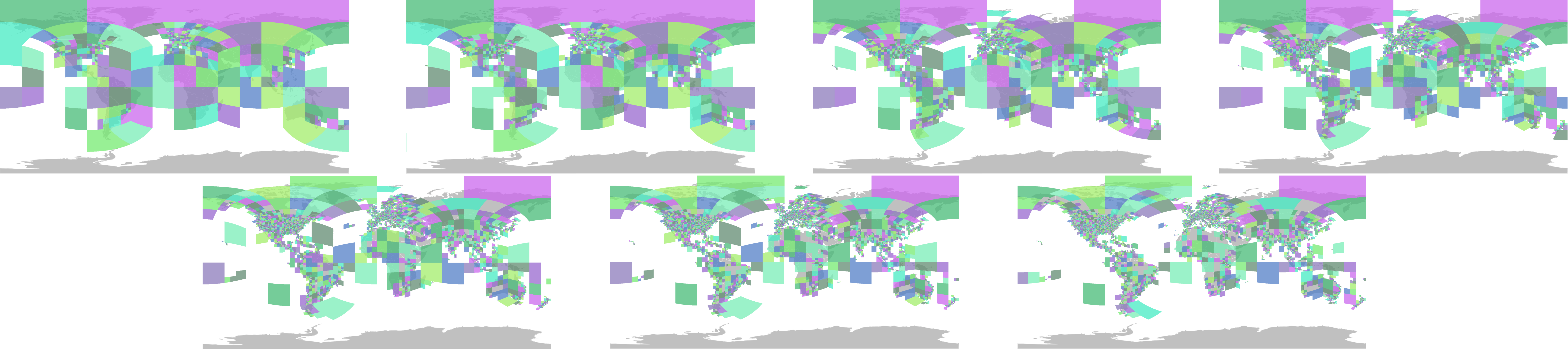}
    \caption{A visualization of all 7 hierarchies used. The $t_{max}$ value is set to 25000, 10000, 5000, 2000, 1000, 750, and  500 respectively for hierarchies 1 to 7, while the $t_{min}$ value is set at 50 for every hierarchy. This generates 684, 1744, 3298, 7202, 12893, 16150, and 21673 classes for hierarchies 1 to 7 respectively.}
    \label{fig:hiers}
\end{figure*}

\section{Introduction}
\label{sec:intro}

Image geo-localization is the task of determining the GPS coordinates of where a photo was taken as precisely as possible. For certain locations, this may be an easy task, as most cities will have noticeable buildings, landmarks, or statues that give away their location. For instance, given an image of the Eiffel Tower one could easily assume it was taken somewhere in Paris. Noticing some of the finer features, like the size of the tower in the image and other buildings that might be visible, a prediction within a few meters could be fairly easy. However, given an image from a small town outside of Paris, it may be very hard to predict its location. Certain trees or a building's architecture may indicate the image is in France, but localizing finer than that can pose a serious challenge. Adding in different times of day, varying weather conditions, and different views of the same location makes this problem even more complex as two images from the same location could look wildly different.

Many works have explored solutions to this problem, with nearly all works focusing on the retrieval task, where query images are matched to a gallery of geo-tagged images to retrieve matching geo-tagged image \cite{regmi2019bridging,shi2019spatial,shi2020looking,toker2021coming,zhu2021vigor,zhu2022transgeo}. There are two variations of the retrieval approach to this problem, same-view and cross-view. In same-view both the query and gallery images are taken at ground level. However, in cross-view the query images are ground level while the gallery images are from an aerial view, either by satellite or drone. This creates a challenging task as images with the exact same location look very different from one another. Regardless of same-view or cross-view, the evaluation of the retrieval task is costly as features need to be extracted and compared for every possible match with geo-tagged gallery images, making global scale geo-localization costly if not infeasible.

If, instead, the problem is approached as a classification task, it's possible to localize on the global scale given enough training data \cite{weyand2016planet,seo2018cplanet,vo2017revisiting,muller2018geolocation,kordopatis2021leveraging,pramanick2022world}. These approaches segment the Earth into Google's S2\footnote{\label{note1}https://code.google.com/archive/p/s2-geometry-library/} cells that are assigned GPS locations and serve as classes, speeding up evaluation. Most previous classification-based visual geo-localization approaches use the same strategy as any other classification task: using an image backbone (either a Convolutional Neural Network or a Vision Transformer \cite{dosovitskiy2020image}), they learn a set image features and output a probability distribution for each possible location (or class) using an MLP. In more recent works \cite{muller2018geolocation,pramanick2022world}, using multiple sets of classes that represent different global scales, as well as utilizing information about the scene characteristics of the image has shown to improve results. These approaches produce one feature vector for an image and presume that it is good enough to localize at every geographic level. However, that is not how a human would reason about finding out their location. If a person had no idea where they were, they would likely search for visual cues for a broad location (country, state) before considering finer areas. Thus, a human would look for a different set of features for each geographic level they want to predict.

In this paper, we introduce a novel approach toward world-wide 
visual geo-localization inspired by human experts. Typically, humans do not evaluate the entirety of a scene and reason about its features, but rather identify important objects, markers, or landmarks and match them to a cache of knowledge about various known locations. In our approach, we emulate this by using a set of learned latent arrays called ``hierarchy queries" that learn a different set of features for each geographic hierarchy. These queries also learn to extract features relative to specific scene types (e.g. forests, sports fields, industrial, etc.). We do this so that our queries can focus more specifically on features relevant to their assigned scene as well as the features related to their assigned hierarchy. This is done via a Transformer Decoder that cross-attends our hierarchy and scene queries with image features that are extracted from a backbone. We also implement a "hierarchy dependent decoder" that ensures our model learns the specifics of each individual hierarchy. To do this our "hierarchy dependent decoder" separates the queries according to their assigned hierarchy, and  has independent weights for the Self-Attention and Feed-Forward stages that are specific to each hierarchy.

\begin{figure}[h]
    \centering
    \includegraphics[width=0.47\textwidth]{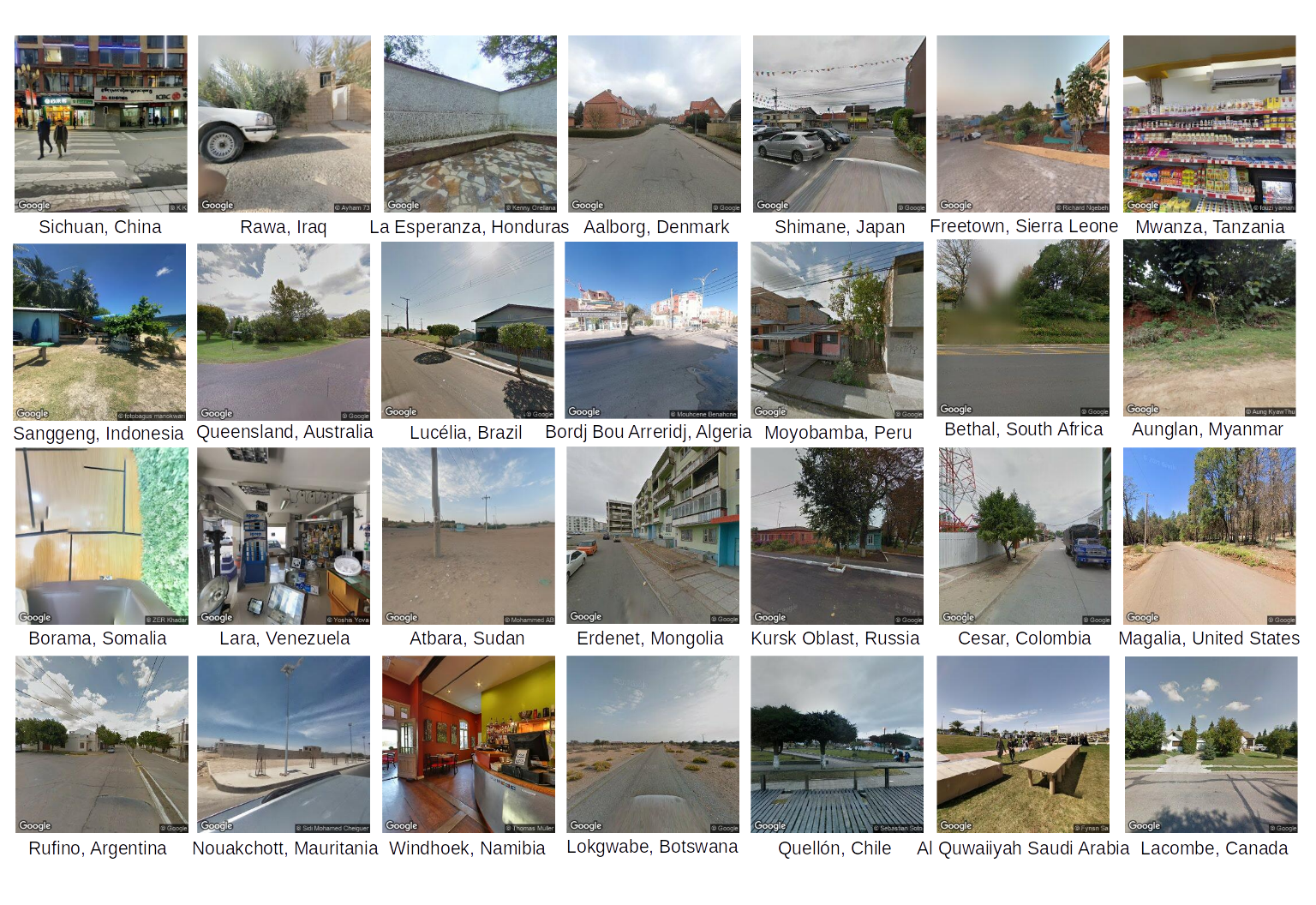}
    \caption{Example images from 28 different countries in the Google-World-Streets-15k dataset}
    \label{fig:GWS15k}
\end{figure}

We also note that the existing testing datasets contain implicit biases which make them unfit to truly measure a model's geo-location accuracy. For instance, Im2GPS \cite{vo2017revisiting, hays2008im2gps} datasets contain many images of iconic landmarks, which only tests whether a model has seen and memorized the locations of those landmarks. Also, YFCC \cite{theiner2022interpretable, vo2017revisiting} testing sets are composed entirely of images posted online that contained geo-tags in their metadata. This creates a bias towards locations that are commonly visited and posted online, like tourist sites. Previous work has found this introduces significant geographical and often racial biases into the datasets \cite{kalkowski2015real} which we demonstrate in Figure \ref{fig:dists}. To this end, we introduce a challenging new testing dataset called Google-World-Streets-15k, which is more evenly distributed across the Earth and consists of real-world images from Google Streetview.

The contributions of our paper include:
(1) The first Transformer Decoder for worldwide image geo-localization.
(2) The first model to produce multiple sets of features for an input image, and the first model capable of extracting scene-specific information without needing a separate network for every scene.
(3) A new testing dataset that reduces landmark bias and reduces biases created by social media.
(4) A significant improvement over previous SOTA methods on all datasets.
(5) A qualitative analysis of the features our model learns for every hierarchy and scene query.

\section{Related Works}
\label{sec:related}
\subsection{Retrieval Based Image Geo-Localization}
The retrieval method for geo-localization attempts to match a query image to target image(s) from a reference database (gallery). Most methods train by using separate models for the ground and aerial views, bringing the features of paired images together in a shared  space. Many different approaches have been proposed to overcome the domain gap, with some methods implementing GANs \cite{goodfellow2014generative} that map images from one view to the other \cite{regmi2019bridging}, others use a polar transform that makes use of the prior geometric knowledge to alter aerial views to look like ground views \cite{shi2019spatial, shi2020looking}, and a few even combine the two techniques in an attempt to have the images appear even more similar \cite{toker2021coming}.

Most methods assume that the ground and aerial images are perfectly aligned spatially. However, this is not always the case. In circumstances where orientation and spatial alignment aren't perfect, the issue can be accounted for ahead of time or even predicted \cite{shi2020looking}. VIGOR \cite{zhu2021vigor} creates a dataset where the spatial location of a query image could be located anywhere within the view of its matching aerial image. Zhu \cite{zhu2022transgeo} strays from the previous methods by using a non-uniform crop that selects the most useful patches of aerial images and ignores others.

\subsection{Image Geo-Localization as Classification}
By segmenting the Earth's surface into distinct classes and assigning a GPS coordinate to each class, a model is allowed to predict a class directly instead of comparing features to a database. Treating geo-localization this way was first introduced by Weyand et al. \cite{weyand2016planet}. In their paper, they introduce a technique to generate classes that utilizes Google's S2 library and a set of training GPS coordinates to partition the Earth into cells, which are treated as classes. Vo \cite{vo2017revisiting} was the first to introduce using multiple different partitions of varying granularity. In contrast, CPlaNet \cite{seo2018cplanet} develops a technique that uses combinatorial partitioning. This approach uses multiple different coarse partitions and encodes each of them as a graph, then refining the graph by merging nodes. More details on class generation will be discussed in Section 3.1.

Up until Individual Scene Networks (ISNs) \cite{muller2018geolocation}, no information other than the image itself was used at training time. The insight behind ISNs was that different image contexts require different features to be learned in order to accurately localize the image. They make use of this by having three separate networks for indoor, natural, and urban images respectively. This way each network can learn the important features for each scene and more accurately predict locations. \cite{muller2018geolocation} also introduced the use of hierarchical classes. While previous papers had utilized multiple geographic partitions, they observed that these partitions could be connected through a hierarchical structure. To make use of this, they proposed a new evaluation technique that combines the predictions of multiple partitions, similar to YOLO9000 \cite{redmon2017yolo9000}, which helps refine the overall prediction. Kordopatis-Zilos \cite{kordopatis2021leveraging} developed a method that combines classification and retrieval. Their network uses classification to get a predicted S2 cell, then retrieval within that cell to get a refined prediction.


Most recently, TransLocator \cite{pramanick2022world} was introduced, which learns from not only the RGB image but also the segmentation map produced by a trained segmentation network. Providing the segmentation map allows TransLocator to rely on the segmentation if there are any variations in the image, like weather or time of day, that would impact a normal RGB-based model. 

All of these methods fail to account for features that are specific to different geographic hierarchies and don't fully utilize scene-specific information. We solve these problems with our query-based learning approach.

\begin{figure*}[h]
    \centering
    \includegraphics[width=1 \textwidth]{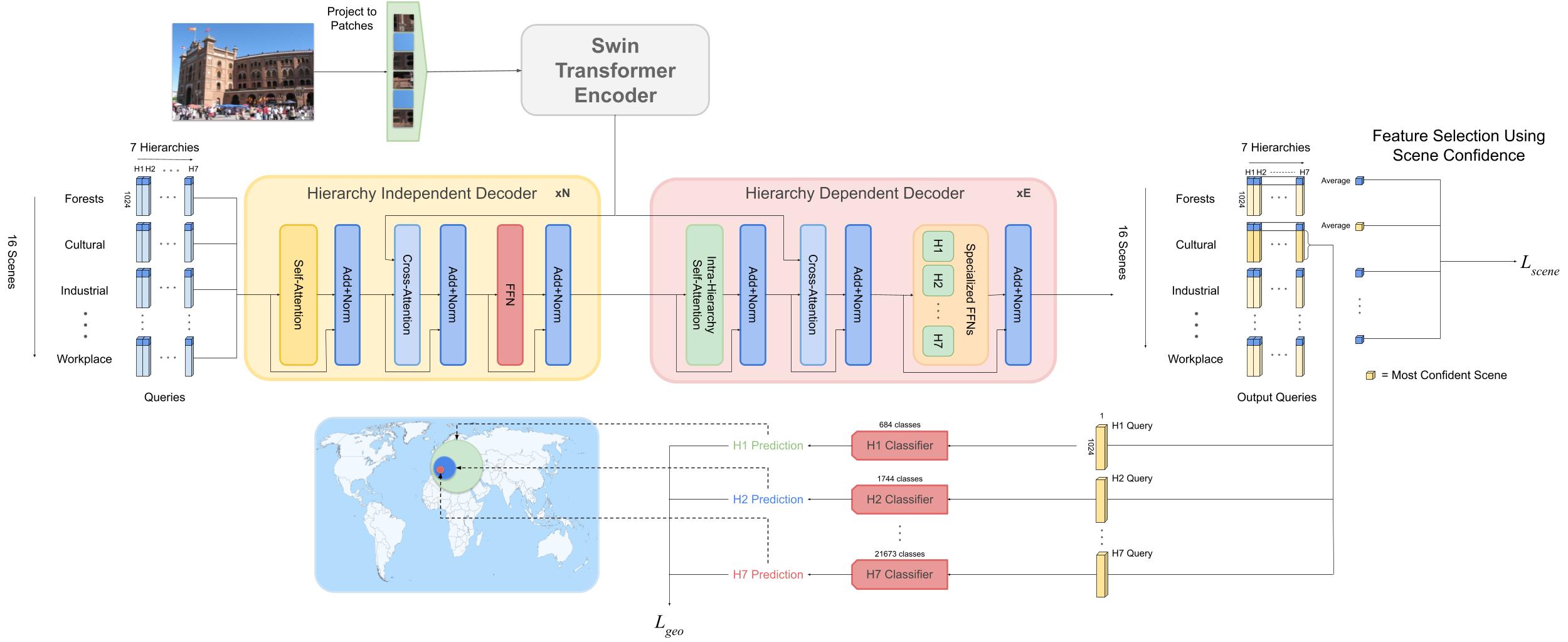}
    \caption{Our proposed network. We randomly initialize a set of learned queries for each hierarchy and scene. An image is first encoded by Transformer Encoder and decoded by two decoders. The first decoder consists of $N$ layers as a Hierarchy Independent Decoder, followed by $E$ layers of our Hierarchy Dependent Decoder; this decoder only performs self-attention within each hierarchy, instead of across all hierarchies, and has separate Feed-Forward Networks for each hierarchy.
    To determine which scene to use for prediction, the scene with the highest average confidence (denoted by the $0^{th}$ channel) is selected and queries are fed to their corresponding classifier 
    to geo-localize at each hierarchy. We get a final prediction by multiplying the class probabilities of the coarser hierarchies into the finer ones so that a prediction using all hierarchical information can be made.}
    \label{fig:arch}
\end{figure*}

\section{Method}
\label{sec:method}
In our approach, we treat discrete locations as classes, obtained by dividing 
the planet into Schneider-2 cells at different levels of geographic granularity. The size of each cell is determined by the number of training images available in the given region, with the constraint that each cell has approximately the same number of samples. We exploit the hierarchical nature of geo-location by learning different sets of features for each geographic hierarchy and for each scene category from an input image.
Finally, we classify a query image by selecting the set of visual features correlated with the most confident scene prediction. We use these sets of features to map the image to an S2 cell at each hierarchical level and combine the predictions at all levels into one refined prediction using the finest hierarchy. 


\subsection{Class Generation}
With global geo-localization comes the problem of separating the Earth into classes. A naive way to do this would be to simply tessellate the earth into  the rectangles that are created by simple latitude and longitude lines. However, this approach has a few issues, for one the surface area of each rectangle will vary with the distance from the poles, likely producing large class imbalances. Instead, we utilize Schneider 2 cells using Google's S2 Library. This process initially projects the Earth's sphere onto the 6 sides of a cube, thereby resulting in an initial 6 S2 cells. To create balanced classes, we split each cell with more than $t_{max}$ images from the training set located inside of it. We also ignore any cells that have less than $t_{min}$ to ensure we don't have classes with an insignificant number of training instances. The cells are split recursively using this criterion until all cells fall within $t_{min}$ and $t_{max}$. This creates a set of balanced classes that cover the entire Earth. These classes and hierarchies are visualized in Figure \ref{fig:hiers} where we can see the increasing specificity of our hierarchies. We begin with 684 classes at our coarsest hierarchy and increase that to 21673 at our finest. During evaluation we define the predicted location as the mean of the location of all training images inside a predicted class.

\subsection{Model}
One problem  faced in geo-localization is that two images in the same geographic cell can share very few visual similarities. Two images from the same location could be taken at night or during the day, in sunny or rainy weather, or simply from the same location but one image faces North while the other faces South. Additionally, some information in a scene can be relevant to one geographic hierarchy (e.g. state) but not another (e.g. country). To that end, we propose a novel decoder-based architecture designed to learn unique sets of features for each of these possible settings. We begin by defining our geographic queries as $GQ \in \mathbb{R}^{HS \times D}$ where $H$ 
is the number of geographic hierarchies,  $S$ is the number of scene labels, and $D$ is the dimension of the features. We  define each individual geographic query as $gq^{h}_{s}$ where $h $ and $s$ represent the index of the hierarchy and scene, respectively. The ground truth scene labels are provided by \textit{Places2} dataset \cite{zhou2017places} which has 365 total scene labels. They also provide a hierarchical structure to their scenes, with coarser hierarchies containing 3 and 16 unique scene labels. A pre-trained scene classification model is used to get the initial scene label from the set of 365 and the coarser labels are extracted using the hierarchical structure. We find that the set of 16 scenes instead of 365 gives the best results for our model, we show ablation on this in supplementary material.

\subsection{GeoDecoder}
We have two decoders as shown in Figure \ref{fig:arch}. Below we describe each in detail.

\noindent{\bf Hierarchy Independent Decoder}\
The geographic queries are passed into our GeoDecoder, whose primary function is, for each hierarchical query, to extract geographical information relevant to its individual task for the image tokens which have been produced by a Swin encoder \cite{liu2021swin}.
As previously stated, our decoder performs operations on a series of learned latent arrays called \textit{Geographic Queries} (scene and hierarchical) in a manner inspired by the Perceiver \cite{jaegle2021perceiver} and DETR \cite{carion2020end}. We define $X$ as the image tokens, $GQ^{k}$ as the geographic queries at the $k^{th}$ layer of the decoder, and $GQ^{0}$ as the initial learned vectors. Each layer performs self-attention on the normalized geographic queries, followed by cross-attention between the output of self-attention and the image patch encodings, where cross-attention is defined as $CA(Q, K) = softmax(\frac{QK^T}{\sqrt{d_k}})K$.
where $Q, K$ are Query and Key respectively.
Finally, we normalize the output of the cross-attention operation and feed it into an MLP to produce the output of the decoder layer. Therefore, one decoder layer is defined as

\begin{align}
    y^{SA} &= MSA(LN(GQ^{k-1})) + GQ^{k-1}\\
    y^{CA} &= CA(LN(y^{SA}, LN(X)) + y^{SA}, \\
    GQ^{k} &= FFN(LN(y^{CA})) + y^{CA}
\end{align}

\noindent{\bf Hierarchy Dependent Decoder}\\
We find that a traditional transformer decoder structure for the entire GeoDecoder results in a homogeneity of all hierarchical queries. Therefore, in the final layers of the decoder, we perform self-attention only in an \textit{intra} hierarchical manner rather than between all hierarchical queries. Additionally, we assign each hierarchy its own fully connected network at the end of each layer rather than allowing hierarchies to share one network. We define the set of geographic queries \textit{specifically} for hierarchy $h$ at layer $k$ $GQ^{k}_{h}$. The feed-forward network for hierarchy $h$ is referred to as $FFN_{h}$
\begin{align}
    y^{SA} &= MSA(LN(GQ^{k-1}_{h})) + GQ^{k-1}_{h}, \\
    y^{CA} &= CA(LN(y^{SA}), LN(X)) + y^{SA}, \\
    GQ^{k}_{h} &= FFN_{h}(LN(y^{CA})) + y^{CA}
\end{align}

After each level, each $GQ^{k}_{h}$ is concatenated to reform $GQ$. In the ablations Table \ref{DecoderDepth-2}, we show the results of these \textit{hierarchy dependent layers}.

\subsection{Losses}
As shown in Figure \ref{fig:arch}, our network is trained with two losses. The first loss is scene prediction loss, $L_{scene}$, which is a Cross-Entropy loss between the predicated scene label $\hat{s_i}$ ground truth scene labels $s_i$. Our second loss is a geo-location prediction loss, $L_{geo}$, which is a combination of Cross-Entropy losses for each hierarchy. Given an image $X$ we define the set of location labels as {$h_1$, $h_2$, ..., $h_7$}, where $h_i$ denotes the ground-truth class distribution in hierarchy $i$, and the respective prediction distribution as $\hat{h_i}$, we define $L_{scene(X)} = CE(s_i, \hat{s_i})$ and $L_{geo}(X) = \sum_{i=1}^7 CE(h_i, \hat{h_i})$ and $L(X) = L_{geo}(X) + L_{scene}(X)$.

\subsection{Inference}
With the output of our GeoDecoder $GQ^{out}$ we can geo-localize the image. As our system is designed to learn different latent embeddings for different visual scenes, we must first choose which features to proceed with. For $gq^{h}_{s} \in GQ$ we assign the confidence that the image belongs to scene $s$ to that vector's $0th$ element. This minimizes the need for an additional individual scene network like in \cite{muller2018geolocation} while allowing specific weights within  the decoder's linear layers to specialize in differentiating visual scenes. Once we have $GQ^{out}$, the queries are separated and sent to the classifier that is assigned to their hierarchy. This gives us 7 different sets of class probabilities, one for each hierarchy. To condense this information into one class prediction, and to exploit the hierarchical nature of our classes, we multiply the probabilities of the classes in the coarser hierarchies by their sub-classes found in the finer hierarchies. If we define a class as $C^{H_i}_j$ where $i$ denotes the hierarchy and $j$ denotes the class label within that hierarchy, we can define the probability of predicting a class $C^{H_7}_a$ for image $X$ as:
 $p(X|C^{H_7}_a)= p(X|C^{H_7}_a)* p(X|C^{H_6}_b)*...* p(X|C^{H_1}_g)$,
given that $C^{H_7}_a$ is a subclass of $C^{H_6}_b$, $C^{H_6}_b$ is a subclass of $C^{H_5}_c$ and so on.
We perform this for every class in our finest hierarchy so we can use the finest geographic granularity while also using the information learned for all of the hierarchies.

\section{Google-World-Streets-15K Dataset}
\label{sec:GWS15}
We propose a new testing dataset collected using Google Streetview called Google-World-Streets-15k. As previous testing datasets contain biases towards commonly visited locations or landmarks, the goal of our dataset is to eliminate those biases and have a more even distribution across the Earth. In total, our dataset contains 14,955 images covering 193 countries.

In order to collect a fair distribution of images, we utilize a database of 43 thousand cities\footnote{https://simplemaps.com/data/world-cities}, as well as the surface area of every country. We first sample a country with a probability proportional to its surface area compared to the Earth's total surface area. Then we select a random city within that country and a GPS coordinate within a 5 Km radius of the center of the city to sample from the Google Streetview API. This ensures that the dataset is evenly distributed according to landmass and not biased towards the countries and locations that people post online.
Google Streetview also blurs out any faces found in the photos, so a model that is using people's faces to predict a location will have to rely on other features in the image to get a prediction.

\begin{figure}
\centering
\begin{subfigure}[b]{0.50\textwidth}
   \includegraphics[width=1\linewidth]{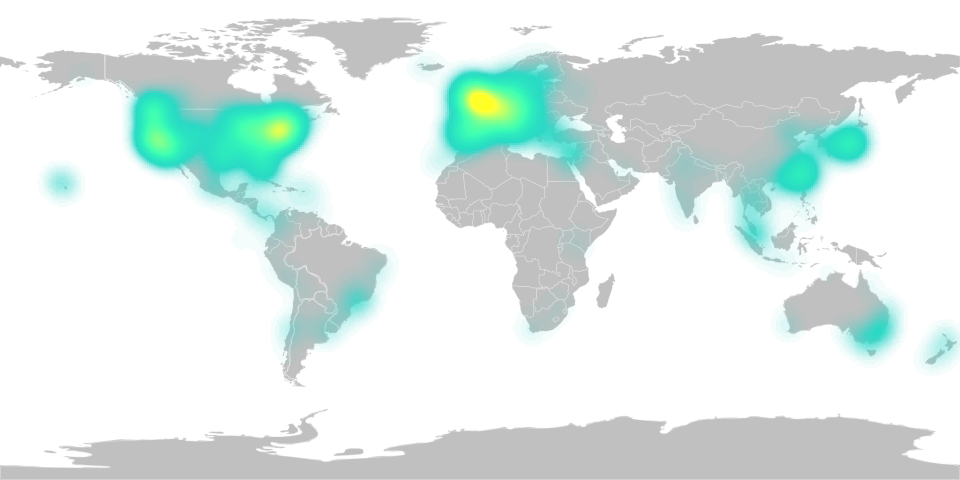}
   \caption{Distribution of images in the YFCC26k validation set}
   \label{fig:Ng1} 
\end{subfigure}

\begin{subfigure}[b]{0.50\textwidth}
   \includegraphics[width=1\linewidth]{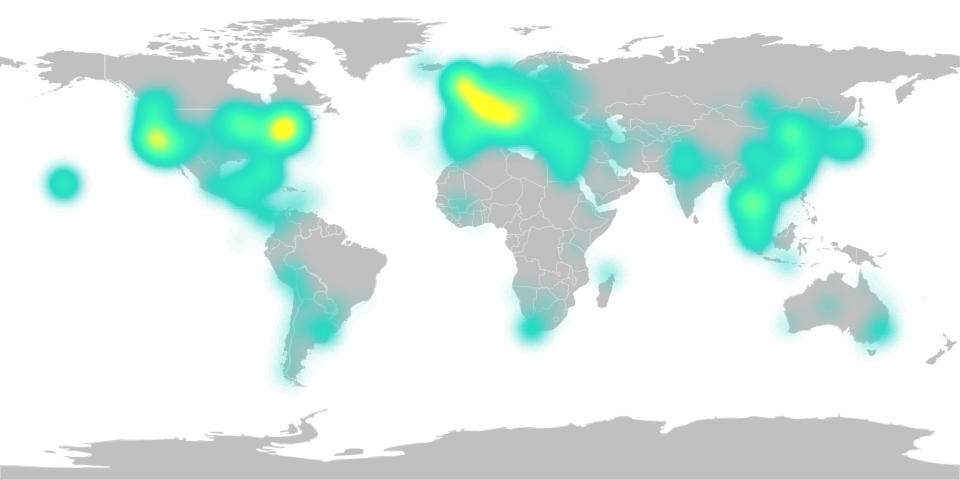}
   \caption{Distribution of images in the Im2GPS3k validation set}
   \label{fig:Ng2} 
\end{subfigure}

\begin{subfigure}[b]{0.50\textwidth}
   \includegraphics[width=1\linewidth]{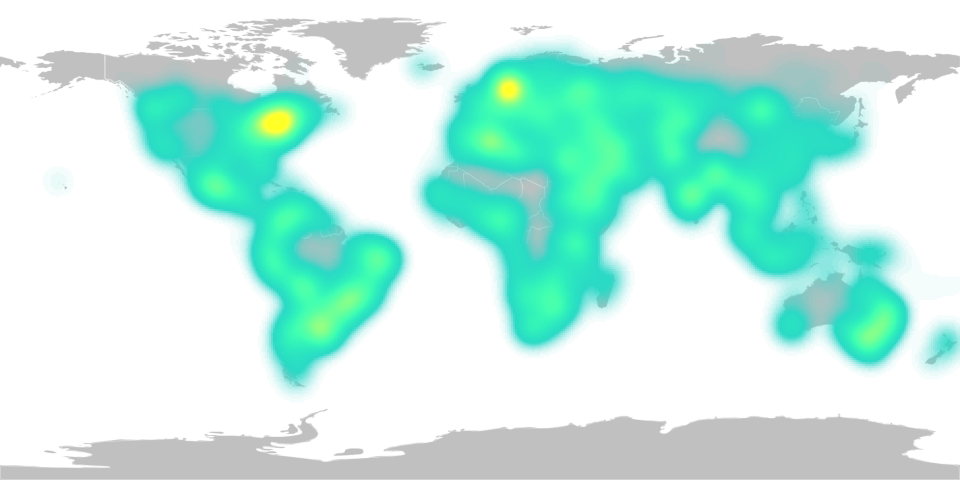}
   \caption{Distribution of images in our Google World Streets 15k validation set}
   \label{fig:Ng3}
\end{subfigure}

\caption[Two numerical solutions]{A comparison of YFCC26k, Im2GPS3k, and our Google World Streets 15k dataset. We see that popular datasets for testing geo-localization systems are heavily concentrated in heavily populated, metropolitan areas, particularly in America and western Europe. By contrast, our dataset more evenly blankets the earth, better representing all countries on earth.}
\label{fig:dists}
\end{figure}

In Figure \ref{fig:dists} We show a heatmap of Google-World-Streets-15k compared to heatmaps of YFCC26k and Im2GPS3k. We note that a majority of YFCC26k and Im2GPS3k are located in North America and Europe, with very little representation in the other 4 populated continents. While Google-World-Streets-15k's densest areas are still the Northeastern US and Europe, we provide a much more even sampling of the Earth with images on all populated continents. We also note that the empty locations on our dataset's heatmap are mostly deserts, tundras, and mountain ranges.

\section{Experiments}
\label{sec:experiments}

\subsection{Training Data}
Our network is trained on the MediaEval Placing Tasks 2016 (MP-16) dataset \cite{larson2017benchmarking}. This dataset consists of 4.72 million randomly chosen geo-tagged images from  the Yahoo Flikr Creative Commons 100 Million (YFCC100M)\cite{thomee2016yfcc100m} dataset. Notably, this subset is fully uncurated, and contains many examples that contain little if any geographic information. These photos include pets, food, and random household objects.  We ensure that no photographer's images appear in both the testing and training sets, to guarantee that our model learns from visual geographic signals rather than the styles of individual photographers. 

\subsection{Testing Data}
We test our method on five datasets: Im2GPS \cite{hays2008im2gps}, Im2GPS3k \cite{vo2017revisiting}, YFCC dataset: YFCC26k \cite{theiner2022interpretable}  YFCC 4k \cite{vo2017revisiting}, and proposed new dataset Google-World-Street-15K described in the previous section.
Im2GPS \cite{hays2008im2gps} and Im2GPS3k \cite{vo2017revisiting}, contain 237 and 2997 images respectively. While small in size, both datasets are manually selected and contain popular sights and landmarks from around the world. We note that many of the landmarks that appear in Im2GPS appear multiple times in the MP-16 dataset, which may cause a bias towards those locations, this is accounted for in our proposed testing dataset.
YFCC dataset: YFCC26k \cite{theiner2022interpretable} and YFCC 4k \cite{vo2017revisiting},  contain 25,600 and 4536 images respectively. In contrast to Im2GPS and like our training set MP-16, these images are randomly selected and often contain very little geo-localizable information, and therefore pose a more difficult challenge than the Im2GPS datasets. 


%
%

\subsection{Evaluation}
During evaluation we utilize the finest hierarchy class to get an image's predicted location. We report our accuracy at the street (1 Km), city (25 Km), region (200 Km), country (750 Km), and continent (2500 Km) scales. However, training on multiple hierarchies allows us to employ a parent-child relationship and multiply 
the probabilities across all hierarchies \cite{muller2018geolocation}.
This allows the finest set of probabilities to be enhanced to include all of the learned hierarchical information. We also use TenCrop during evaluation, which is a cropping technique that returns the four corner crops, center crop, and their flipped versions. All crops are passed through the model and their outputs are averaged to get one set of probabilities per hierarchy for each image.

\begin{table}[t!]
\centering 
\caption{\textbf{Geo-localization accuracy of our proposed method compared to previous methods, across four baseline datasets, and our proposed dataset.} Results denoted with * are using our recreation of the given model.} 

\resizebox{\columnwidth}{!}
{
\begin{tabular}{c | c || c c c c c}
\hline
\multirow{3}{*}{\bf Dataset} & \multirow{3}{*}{\bf \centering Method} & \multicolumn{5}{c}{\bf Distance ($a_r$ [\%] @ km)} \\ 

& & \multirow{1}{1.4 cm}{\tt \bf \centering Street} & \multirow{1}{1.4 cm}{\tt \bf \centering City} & \multirow{1}{1.4 cm}{\tt \bf \centering Region} & \multirow{1}{1.4 cm}{\tt \bf \centering Country} & \multirow{1}{1.4 cm}{\tt \bf \centering Continent} \\ 

& & \bf $1$ km & \bf $25$ km & \bf $200$ km & \bf $750$ km & \bf $2500$ km \\

\hline
\hline

\multirow{10}{1.2 cm}{\tt \bf \centering Im$2$GPS\\~\cite{hays2008im2gps}} & \centering Human \cite{vo2017revisiting} & $-$ & $-$ & 3.8 & 13.9 & 39.3 \\



& \centering [L]kNN, $\sigma$ = $4$ \cite{vo2017revisiting} & 14.4 & 33.3 & 47.7 & 61.6 & 73.4 \\

& \centering MvMF \cite{izbicki2019exploiting} & 8.4 & 32.6 & 39.4 & 57.2 & 80.2\\

& \centering PlaNet \cite{weyand2016planet}  & 8.4 & 24.5 & 37.6 & 53.6 & 71.3\\

& \centering CPlaNet \cite{seo2018cplanet}  & 16.5 & 37.1 & 46.4 & 62.0 & 78.5\\

& \centering ISNs (M, f, S$_3$) \cite{muller2018geolocation} & 16.5 & 42.2 & 51.9 & 66.2 & 81.0 \\

& \centering ISNs (M,f$^*$,S$_3$) \cite{muller2018geolocation} & 16.9 & 43.0 & 51.9 & 66.7 & 80.2 \\  


& Translocator & 19.9 & 48.1 & 64.6 & 75.6 & 86.7 \\


& Ours & \bf 22.1 & \bf 50.2 & \bf 69.0 & \bf 80.0 & \bf 89.1\\

\hline
\hline

\multirow{8}{1.2 cm}{\tt \centering \bf Im$2$GPS\\$3$k\\ \cite{vo2017revisiting}} & [L]kNN, $\sigma$ = $4$ \cite{vo2017revisiting} & 7.2 & 19.4 & 26.9 & 38.9 & 55.9 \\

& PlaNet$^\dagger$ \cite{weyand2016planet} & 8.5 & 24.8 & 34.3 & 48.4 & 64.6 \\

& CPlaNet \cite{seo2018cplanet} & 10.2 & 26.5 & 34.6 & 48.6 & 64.6 \\

& ISNs (M, f, S$_3$) \cite{muller2018geolocation} & 10.1 & 27.2 & 36.2 & 49.3 & 65.6 \\

& ISNs (M,f$^*$,S$_3$) \cite{muller2018geolocation} & 10.5 & 28.0 & 36.6 & 49.7 & 66.0 \\  


& Translocator &  11.8 & 31.1 & \bf 46.7 & 58.9 & \bf 80.1 \\

& Ours & {\textbf{12.8}}  & {\textbf{33.5}} & {45.9}  & {\textbf{61.0}} & {76.1} \\


\hline 
\hline

\multirow{9}{1.2 cm}{\tt \centering \bf YFCC\\$4k$\\ \cite{vo2017revisiting}} & [L]kNN, $\sigma$ = $4$ \cite{vo2017revisiting} & 2.3 & 5.7 & 11.0 & 23.5 & 42.0 \\

& PlaNet$^\dagger$ \cite{weyand2016planet} & 5.6 & 14.3 & 22.2 & 36.4 & 55.8 \\

& CPlaNet \cite{seo2018cplanet} & 7.9 & 14.8 & 21.9 & 36.4 & 55.5\\

& ISNs (M, f, S$_3$)$^\ddagger$ \cite{muller2018geolocation} & 6.5 & 16.2 & 23.8 & 37.4 & 55.0 \\

& ISNs (M,f$^*$,S$_3$)$^\ddagger$ \cite{muller2018geolocation} & 6.7 & 16.5 & 24.2 & 37.5 & 54.9 \\  


& Translocator & 8.4 & 18.6 & 27.0 & 41.1  & 60.4 \\


& Ours & {\textbf{10.3}}  & {\textbf{24.4}} & {\textbf{33.9}}  & {\textbf{50.0}} & {\textbf{68.7}}\\

\hline 
\hline

\multirow{6}{1.2 cm}{\tt \centering \bf YFCC\\$26$k\\~\cite{theiner2022interpretable}}& PlaNet$^\ddagger$ \cite{weyand2016planet} & 4.4 & 11.0 & 16.9 & 28.5 & 47.7\\

& ISNs (M, f, S$_3$)$^\ddagger$ \cite{muller2018geolocation} & 5.3 & 12.1 & 18.8 & 31.8 & 50.6 \\

& ISNs (M, f$^*$, S$_3$)$^\ddagger$ \cite{muller2018geolocation} &  5.3 & 12.3 & 19.0 & 31.9 & 50.7 \\ 


& Translocator & 7.2 & 17.8 & 28.0 & 41.3 & 60.6\\


& Ours & {\textbf{10.1}}  & {\textbf{23.9}} & {\textbf{34.1}}  & {\textbf{49.6}} & {\textbf{69.0}}\\

\hline 
\hline

\multirow{3}{1.2 cm}{\tt \centering \bf GWS\\$15$k}

& ISNs (M, f$^*$, S$_3$)$^\ddagger$ \cite{muller2018geolocation} &  0.05 & 0.6 & 4.2 & 15.5 & 38.5 \\

& Translocator* & 0.5 & 1.1 & 8.0 & 25.5 & 48.3\\


& Ours & {\textbf{0.7}}  & {\textbf{1.5}} & {\textbf{8.7}}  & {\textbf{26.9}} & {\textbf{50.5}}\\

\hline
\hline
\end{tabular}}

\label{tab:results_main}
\vspace{-3mm}
\end{table}

\vspace{-2mm}

\section{Results, Discussions and Analysis}

In this section, we compare the performance of our method with different baselines, and conduct a detailed ablation study to demonstrate the importance of different components in our system. Furthermore, we visualize the interpretability of our method by showing the attention map between each query and the image patches from our encoder. 

Our results are presented in Table \ref{tab:results_main}. On Im2GPS, our method achieves state of the art accuracy across all distances, improving by as much as 1.7\% on the baseline. For Im2GPS3k our method manages to beat the previous techniques on a majority of distances, only falling short on the 200 and 2500 kilometer accuracies. More notably, our system's performance on the far more challenging YFCC4k and YFCC26k datasets vastly outperforms previous geo-localization works. On YFCC4k, 
ours achieves a score of 10.3\%, an improvement of 2.2\% over Translocator. Similarly on YFCC26k, we achieve a 1KM accuracy of 10.1\%, improving over 
Translocator by 
2.9\%. Additionally, we compared our method to \cite{pramanick2022world} on our Google-World-Streets-15k(GWS) validation dataset. As expected, the more realistic and fair nature of this dataset, in contrast to the training set MP-16, resulted in poor performance on all systems. However, we still outperform Translocator by 0.2\% on 1KM accuracy and 0.4\% on 25KM accuracy, suggesting a stronger ability to focus on defining features of a scene, rather than singular landmarks.
\subsection{Qualitative Results}
\begin{figure*}[h]
    \centering
    \includegraphics[width=17cm]{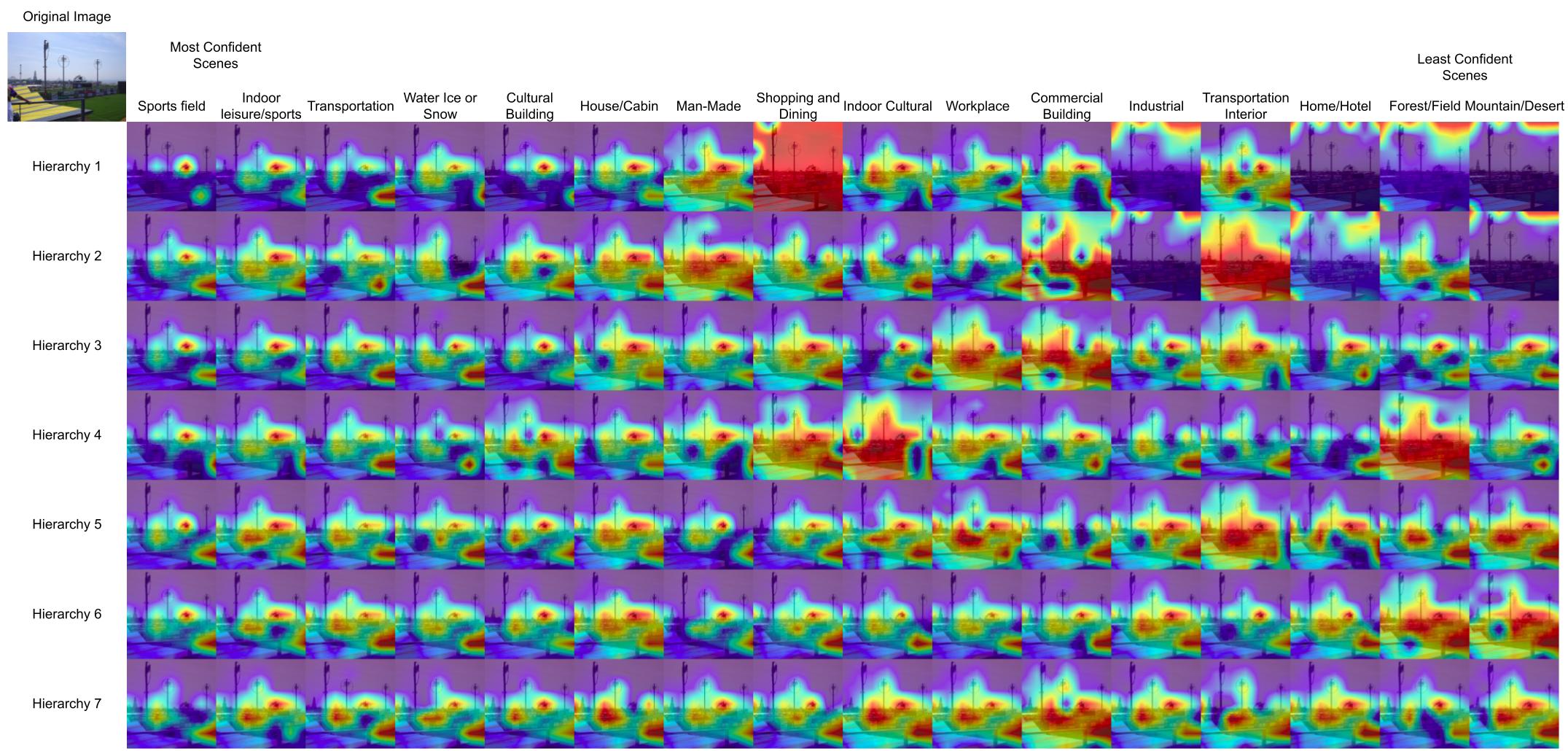}
    \caption{A qualitative analysis of different queries. Here we show the attention maps between every query our model produces when probed with the original Im2GPS3k image seen in the top left. Each row shows a hierarchy query for all scenes, while each column shows each scene query for all hierarchies. This specific query image is of an outdoor sports field. We observe that the most relevant scene labels were predicted as most confident and that their attention maps are more localized to specific features that would define a sports field. Looking at the less confident scenes, we see that the attention maps look at more general features or at random areas of the image. This is because those queries are trained to find features for their specific scenes. For example, the shopping and dining query will be looking for things like tables, chairs, or storefronts that aren't present in this query image, which is why we see the attention maps looking more generally at the image rather than looking at specific features.}
    \label{fig:scene_attn}
\end{figure*}

We provide a number of qualitative results, outlined in Figure \ref{fig:scene_attn}. For our attention maps, we use the attention between the image backbone features and the fine-level query (corresponding to the correct scene). First, these results show that each hierarchy query attends to different parts of the image, as per our original hypothesis. Second, we can see that the attention for the \textit{correct} scene query is far more precise than \textit{incorrect} scene queries, demonstrating how our system learns different features specific to each scene.

\subsection{Ablations}

\begin{table}[t!]
\centering
\caption{\textbf{Ablation Study on GeoDecoder Depth} We find that larger depths offer marginal increases in performance, and there are diminishing returns for more than 8 layers.}
\resizebox{\columnwidth}{!}
{
\begin{tabular}{c | c || c c c c c}
\hline
\multirow{3}{*}{\textbf{Dataset}} & \multirow{3}{*}{\textbf{Depth}} & \multicolumn{5}{c}{\textbf{Distance ($a_r$ [\%] @ km)}} \\

& & \multirow{1}{1.4 cm}{\tt \bf \centering Street} & \multirow{1}{1.4 cm}{\tt \bf \centering City} & \multirow{1}{1.4 cm}{\tt \bf \centering Region} & \multirow{1}{1.4 cm}{\tt \bf \centering Country} & \multirow{1}{1.4 cm}{\tt \bf \centering Continent} \\

& & \bf $1$ km & \bf $25$ km & \bf $200$ km & \bf $750$ km & \bf $2500$ km \\

\hline
\hline

 & 3 & 11.9 & 32.9 & 45.0 & 59.5 & 75.4 \\
 
\tt \bf \centering Im2GPS3k & 5 & 12.5 & 33.3 & 45.2 & 60.1 & 75.9 \\

\cite{vo2017revisiting}& 8 & \textbf{12.8} & \textbf{33.5} & \textbf{45.9} & \textbf{61.0} & 76.1 \\

& 10 & 12.5 & 33.2 & 45.2 & 60.1 & \textbf{76.2} \\

\hline
\hline

& 3 & 9.7 & 23.5 & 33.4 & 49.3 & 68.3 \\

\tt \bf \centering YFCC26k& 5 & 9.9 & 23.6 & 33.8 & 49.6 & 68.5 \\ 

\cite{theiner2022interpretable}& 8 & \textbf{10.1} & \textbf{23.9} & \textbf{34.1} & 49.6 & 69.0 \\

& 10 & 10.0 & 23.7 & 33.6 & \textbf{50.1} & \textbf{69.2}\\

\hline
\hline
\end{tabular}
}
\label{DecoderDepth-1}
\end{table}

\begin{table}[t!]
\centering 
\caption{\textbf{Ablation Study on scene prediction method} We show our max score selection method of scene queries outperforms both  scene prediction  approach of \cite{pramanick2022world}, as well as treating scenes as an additional task.} 

\resizebox{\columnwidth}{!}
{
\begin{tabular}{c | c || c c c c c}
\hline
\multirow{3}{*}{\bf Dataset} & \multirow{3}{*}{\bf \centering Method} & \multicolumn{5}{c}{\bf Distance ($a_r$ [\%] @ km)} \\ 

& & \multirow{1}{1.4 cm}{\tt \bf \centering Street} & \multirow{1}{1.4 cm}{\tt \bf \centering City} & \multirow{1}{1.4 cm}{\tt \bf \centering Region} & \multirow{1}{1.4 cm}{\tt \bf \centering Country} & \multirow{1}{1.4 cm}{\tt \bf \centering Continent} \\ 

& & \bf $1$ km & \bf $25$ km & \bf $200$ km & \bf $750$ km & \bf $2500$ km \\

\hline
\hline

\multirow{3}{1.6 cm}{\tt \bf \centering Im2GPS3k\\ \cite{vo2017revisiting}} 

& No Scene Prediction & 11.7 & 31.5 & 42.3 & 57.0 & 72.3 \\ 
& Scene Prediction \cite{pramanick2022world} & 12.2 & 32.8 & 44.3 & 59.5 & 75.8 \\ 

& Ours & \textbf{12.8} & \textbf{33.5} & \textbf{45.9} & \textbf{61.0} & \textbf{76.1} \\

\hline
\hline

\multirow{3}{1.6 cm}{\tt \bf \centering YFCC26k \\ \cite{theiner2022interpretable}}  

& No Scene Prediction & 9.4 & 22.9 & 32.6 & 48.0 & 65.4 \\ 
& Scene Prediction \cite{pramanick2022world} & 9.7 & 23.2 & 33.0 & 48.8 & 67.0 \\ 

& Ours & \textbf{10.1} & \textbf{23.9} & \textbf{34.1} & \textbf{49.6} & \textbf{69.0} \\

\hline 
\hline
\end{tabular}}

\label{ScenePrediction}
\vspace{-0.2cm}
\end{table}

\textbf{Ablation Study on Encoder Type}
We performed an ablations study on different image encoders. We show that our method outperforms using ViT or Swin on their own. See Table \ref{EncoderType}.


\begin{table}[t!]
\centering 
\caption{\textbf{Ablation Study on Hierarchy Dependent Decoder 
} We show that converting the final two layers of the GeoDecoder to be hierarchy dependent layers offers marginal returns. } 

\resizebox{\columnwidth}{!}
{
\begin{tabular}{c | c || c c c c c}
\hline
\multirow{3}{*}{\bf Dataset} & \multirow{3}{*}{\bf \centering Layers} & \multicolumn{5}{c}{\bf Distance ($a_r$ [\%] @ km)} \\ 

& & \multirow{1}{1.4 cm}{\tt \bf \centering Street} & \multirow{1}{1.4 cm}{\tt \bf \centering City} & \multirow{1}{1.4 cm}{\tt \bf \centering Region} & \multirow{1}{1.4 cm}{\tt \bf \centering Country} & \multirow{1}{1.4 cm}{\tt \bf \centering Continent} \\ 

& & \bf $1$ km & \bf $25$ km & \bf $200$ km & \bf $750$ km & \bf $2500$ km \\

\hline
\hline

\multirow{4}{1.2 cm}{\tt \bf \centering Im2GPS\\3k\\ \cite{vo2017revisiting}}

& 0 & 12.2 & 33.2 & 45.5 & 60.3 & 75.8 \\

& 2 & \textbf{12.8} & \textbf{33.5} & \textbf{45.9} & \textbf{61.0} & \textbf{76.1} \\ 

& 4 & 12.8 & 33.4 & 45.0 & 60.7 & 75.6 \\

& 6 & 12.6 & 33.2 & 44.5 & 59.9 & 75.3 \\

\hline
\hline

\multirow{4}{1.4 cm}{\tt \bf \centering YFCC26k \\ \cite{theiner2022interpretable}}  

& 0 & 9.7 & 23.5 & 33.8 & 49.2 & 68.7 \\

& 2 & \textbf{10.1} & \textbf{23.9} & \textbf{34.1} & \textbf{49.6} & \textbf{69.0} \\ 

& 4 & 9.9 & 23.4 & 33.6 & 49.0 & 68.3 \\

& 6 & 8.7 & 22.6 & 33.0 & 48.6 & 67.6\\

\hline 
\hline
\end{tabular}}
\vspace{0.1cm}

\label{DecoderDepth-2}
\vspace{-6mm}
\vspace{0.2cm}
\end{table}

\begin{table}[t!]
\centering 
\caption{\textbf{Ablation Study on Encoder Type} We show our method performs better than simple image encoders.}

\resizebox{\columnwidth}{!}
{
\begin{tabular}{c | c || c c c c c}
\hline
\multirow{4}{*}{\bf Dataset} & \multirow{4}{*}{\bf Model} & \multicolumn{5}{c}{\bf Distance ($a_r$ [\%] @ km)} \\ 

& & \multirow{1}{1.4 cm}{\tt \bf \centering Street} & \multirow{1}{1.4 cm}{\tt \bf \centering City} & \multirow{1}{1.4 cm}{\tt \bf \centering Region} & \multirow{1}{1.4 cm}{\tt \bf \centering Country} & \multirow{1}{1.4 cm}{\tt \bf \centering Continent} \\ 

& & \bf $1$ km & \bf $25$ km & \bf $200$ km & \bf $750$ km & \bf $2500$ km \\

\hline
\hline

& ViT & 6.9 & 17.3 & 27.5 & 40.5 & 59.5 \\
\bf \centering{YFCC26k} & Swin & 9.6 & 22.3 & 33.6 & 48.0 & 67.5 \\
\cite{theiner2022interpretable}& Ours (ViT) & 8.7 & 21.4 & 31.6 & 47.8 & 66.2 \\
& Ours (Swin) & \textbf{10.1} & \textbf{23.9} & \textbf{34.1} & \textbf{49.6} & \textbf{69.0} \\

\hline
\hline
\end{tabular}}

\label{EncoderType}
\vspace{-0.2cm}
\end{table}

\textbf{GeoDecoder Depth} We perform two ablations on the architecture of the GeoDecoder. First, we experiment with the GeoDecoder's depth, varying it at $n = {3, 5, 8, 10}$ (Table \ref{DecoderDepth-1}). We see a steady improvement from 3 through 8, but then a clear reduction in performance on all metrics at $n=10$. This suggests a point of diminishing returns. Additionally, we experiment with the hierarchy dependent layers on the end of the GeoDecoder (Table \ref{DecoderDepth-2}). Recall, these layers restrict attention operations to queries within the same hierarchy, as well as utilize specialized feed-forward layers.  For these experiments we vary the number of hierarchy dependent layers, while assigning the remaining layers to the hierarchy independent decoder, such that the total number of layers is always 8.

\textbf{Scene Prediction} One contribution of our method is our approach toward distinguishing between different visual scenes of the same location. To show the effectiveness of our separated scene queries, we ablate on scene prediction by evaluating performance with no scene prediction, as well as using scene prediction as a secondary task as in \cite{pramanick2022world}. We then compare it to our scene prediction method. See (Table \ref{ScenePrediction}). Our results find that our scene queries selection method outperforms treating scenes as a secondary task by 0.6\% and 0.4\% on Im2GPS3k and YFCC26k, respectively. 

\textbf{Additional Ablations} We perform additional ablations on the number of scenes as well as the number of hierarchies in the supplementary.

\section{Conclusion}
\label{sec:conclusion}
In this work, we reformulated visual geo-localization via the learning of multiple sets of geographic features. Given an RGB image of any location on planet earth, our system first learns a set of patch features, then uses the GeoDecoder to extract hierarchy-specific features for each possible scene, choosing the most confident scene before prediction. Our proposed method improves over other geo-localization methods on multiple benchmarks, especially on uncurated datasets most similar to real-world use cases.

\clearpage
{\small
\bibliographystyle{ieee_fullname}
\bibliography{egbib}

\begin{thebibliography}{10}\itemsep=-1pt

\bibitem{carion2020end}
Nicolas Carion, Francisco Massa, Gabriel Synnaeve, Nicolas Usunier, Alexander
  Kirillov, and Sergey Zagoruyko.
\newblock End-to-end object detection with transformers.
\newblock In {\em European conference on computer vision}, pages 213--229.
  Springer, 2020.

\bibitem{dosovitskiy2020image}
Alexey Dosovitskiy, Lucas Beyer, Alexander Kolesnikov, Dirk Weissenborn,
  Xiaohua Zhai, Thomas Unterthiner, Mostafa Dehghani, Matthias Minderer, Georg
  Heigold, Sylvain Gelly, et~al.
\newblock An image is worth 16x16 words: Transformers for image recognition at
  scale.
\newblock {\em arXiv preprint arXiv:2010.11929}, 2020.

\bibitem{gates1909mutability}
RR Gates.
\newblock Mutability and variability.
\newblock {\em Botanical Gazette}, 47(6):476--477, 1909.

\bibitem{goodfellow2014generative}
Ian Goodfellow, Jean Pouget-Abadie, Mehdi Mirza, Bing Xu, David Warde-Farley,
  Sherjil Ozair, Aaron Courville, and Yoshua Bengio.
\newblock Generative adversarial nets.
\newblock {\em Advances in neural information processing systems}, 27, 2014.

\bibitem{hays2008im2gps}
James Hays and Alexei~A Efros.
\newblock Im2gps: estimating geographic information from a single image.
\newblock In {\em 2008 ieee conference on computer vision and pattern
  recognition}, pages 1--8. IEEE, 2008.

\bibitem{izbicki2019exploiting}
Mike Izbicki, Evangelos~E Papalexakis, and Vassilis~J Tsotras.
\newblock Exploiting the earth’s spherical geometry to geolocate images.
\newblock In {\em Joint European Conference on Machine Learning and Knowledge
  Discovery in Databases}, pages 3--19. Springer, 2019.

\bibitem{jaegle2021perceiver}
Andrew Jaegle, Felix Gimeno, Andy Brock, Oriol Vinyals, Andrew Zisserman, and
  Joao Carreira.
\newblock Perceiver: General perception with iterative attention.
\newblock In {\em International conference on machine learning}, pages
  4651--4664. PMLR, 2021.

\bibitem{kalkowski2015real}
Sebastian Kalkowski, Christian Schulze, Andreas Dengel, and Damian Borth.
\newblock Real-time analysis and visualization of the yfcc100m dataset.
\newblock In {\em Proceedings of the 2015 workshop on community-organized
  multimodal mining: opportunities for novel solutions}, pages 25--30, 2015.

\bibitem{kordopatis2021leveraging}
Giorgos Kordopatis-Zilos, Panagiotis Galopoulos, Symeon Papadopoulos, and
  Ioannis Kompatsiaris.
\newblock Leveraging efficientnet and contrastive learning for accurate
  global-scale location estimation.
\newblock In {\em Proceedings of the 2021 International Conference on
  Multimedia Retrieval}, pages 155--163, 2021.

\bibitem{larson2017benchmarking}
Martha Larson, Mohammad Soleymani, Guillaume Gravier, Bogdan Ionescu, and
  Gareth~JF Jones.
\newblock The benchmarking initiative for multimedia evaluation: Mediaeval
  2016.
\newblock {\em IEEE MultiMedia}, 24(1):93--96, 2017.

\bibitem{liu2021swin}
Ze Liu, Yutong Lin, Yue Cao, Han Hu, Yixuan Wei, Zheng Zhang, Stephen Lin, and
  Baining Guo.
\newblock Swin transformer: Hierarchical vision transformer using shifted
  windows.
\newblock In {\em Proceedings of the IEEE/CVF International Conference on
  Computer Vision}, pages 10012--10022, 2021.

\bibitem{lorenz1905methods}
Max~O Lorenz.
\newblock Methods of measuring the concentration of wealth.
\newblock {\em Publications of the American statistical association},
  9(70):209--219, 1905.

\bibitem{muller2018geolocation}
Eric Muller-Budack, Kader Pustu-Iren, and Ralph Ewerth.
\newblock Geolocation estimation of photos using a hierarchical model and scene
  classification.
\newblock In {\em Proceedings of the European Conference on Computer Vision
  (ECCV)}, pages 563--579, 2018.

\bibitem{pramanick2022world}
Shraman Pramanick, Ewa~M Nowara, Joshua Gleason, Carlos~D Castillo, and Rama
  Chellappa.
\newblock Where in the world is this image? transformer-based geo-localization
  in the wild.
\newblock {\em arXiv preprint arXiv:2204.13861}, 2022.

\bibitem{redmon2017yolo9000}
Joseph Redmon and Ali Farhadi.
\newblock Yolo9000: better, faster, stronger.
\newblock In {\em Proceedings of the IEEE conference on computer vision and
  pattern recognition}, pages 7263--7271, 2017.

\bibitem{regmi2019bridging}
Krishna Regmi and Mubarak Shah.
\newblock Bridging the domain gap for ground-to-aerial image matching.
\newblock In {\em Proceedings of the IEEE/CVF International Conference on
  Computer Vision}, pages 470--479, 2019.

\bibitem{ridnik2021imagenet}
Tal Ridnik, Emanuel Ben-Baruch, Asaf Noy, and Lihi Zelnik-Manor.
\newblock Imagenet-21k pretraining for the masses.
\newblock {\em arXiv preprint arXiv:2104.10972}, 2021.

\bibitem{seo2018cplanet}
Paul~Hongsuck Seo, Tobias Weyand, Jack Sim, and Bohyung Han.
\newblock Cplanet: Enhancing image geolocalization by combinatorial
  partitioning of maps.
\newblock In {\em Proceedings of the European Conference on Computer Vision
  (ECCV)}, pages 536--551, 2018.

\bibitem{shi2019spatial}
Yujiao Shi, Liu Liu, Xin Yu, and Hongdong Li.
\newblock Spatial-aware feature aggregation for image based cross-view
  geo-localization.
\newblock {\em Advances in Neural Information Processing Systems}, 32, 2019.

\bibitem{shi2020looking}
Yujiao Shi, Xin Yu, Dylan Campbell, and Hongdong Li.
\newblock Where am i looking at? joint location and orientation estimation by
  cross-view matching.
\newblock In {\em Proceedings of the IEEE/CVF Conference on Computer Vision and
  Pattern Recognition}, pages 4064--4072, 2020.

\bibitem{theiner2022interpretable}
Jonas Theiner, Eric M{\"u}ller-Budack, and Ralph Ewerth.
\newblock Interpretable semantic photo geolocation.
\newblock In {\em Proceedings of the IEEE/CVF Winter Conference on Applications
  of Computer Vision}, pages 750--760, 2022.

\bibitem{thomee2016yfcc100m}
Bart Thomee, David~A Shamma, Gerald Friedland, Benjamin Elizalde, Karl Ni,
  Douglas Poland, Damian Borth, and Li-Jia Li.
\newblock Yfcc100m: The new data in multimedia research.
\newblock {\em Communications of the ACM}, 59(2):64--73, 2016.

\bibitem{toker2021coming}
Aysim Toker, Qunjie Zhou, Maxim Maximov, and Laura Leal-Taix{\'e}.
\newblock Coming down to earth: Satellite-to-street view synthesis for
  geo-localization.
\newblock In {\em Proceedings of the IEEE/CVF Conference on Computer Vision and
  Pattern Recognition}, pages 6488--6497, 2021.

\bibitem{vo2017revisiting}
Nam Vo, Nathan Jacobs, and James Hays.
\newblock Revisiting im2gps in the deep learning era.
\newblock In {\em Proceedings of the IEEE international conference on computer
  vision}, pages 2621--2630, 2017.

\bibitem{weyand2016planet}
Tobias Weyand, Ilya Kostrikov, and James Philbin.
\newblock Planet-photo geolocation with convolutional neural networks.
\newblock In {\em European Conference on Computer Vision}, pages 37--55.
  Springer, 2016.

\bibitem{zhou2017places}
Bolei Zhou, Agata Lapedriza, Aditya Khosla, Aude Oliva, and Antonio Torralba.
\newblock Places: A 10 million image database for scene recognition.
\newblock {\em IEEE transactions on pattern analysis and machine intelligence},
  40(6):1452--1464, 2017.

\bibitem{zhu2022transgeo}
Sijie Zhu, Mubarak Shah, and Chen Chen.
\newblock Transgeo: Transformer is all you need for cross-view image
  geo-localization.
\newblock {\em arXiv preprint arXiv:2204.00097}, 2022.

\bibitem{zhu2021vigor}
Sijie Zhu, Taojiannan Yang, and Chen Chen.
\newblock Vigor: Cross-view image geo-localization beyond one-to-one retrieval.
\newblock In {\em Proceedings of the IEEE/CVF Conference on Computer Vision and
  Pattern Recognition}, pages 3640--3649, 2021.

\end{thebibliography}
}

\clearpage

\section{Supplementary}
This article forms the supplementary material for our paper which aims to provide a better insight into our methods, and also provide additional details which we were unable to include. Here, we expand upon the following :
\begin{enumerate}
    \item Additional Ablation Studies
    \begin{enumerate}
        \item Ablation on Number of Scenes
        \item Ablation on Number of Hierarchies
    \end{enumerate}
    \item Analysis of the Google World Streets 15k dataset
    \begin{enumerate}
        \item Lorenz Curves
        \item Gini Coefficient
        \item Examples of Image Localizations
    \end{enumerate}
    \item Review of Baselines
    \begin{enumerate}
        \item Encoder Baselines
        \item Pre-existing Methods
    \end{enumerate}
    \item Implementation Details
    \begin{enumerate}
        \item Hyperparameters
        \item Augmentations
    \end{enumerate}
    \item Qualitative Analysis
\end{enumerate}

\subsection{Additional Ablation Studies}
\subsubsection{Ablation on Number of Scenes} 
Previous works \cite{muller2018geolocation} \cite{pramanick2022world} have already emphasized the importance of having scene-type information in training labels for geo-localization, so this was not the focus of our work. However, we provide results to both show the necessity of these labels, as well as the optimal number to use. On both IM2GPS3k \cite{vo2017revisiting} as well as YFCC26k \cite{theiner2022interpretable}, we found diminishing returns when using 365, with a drop of 0.2\% 1KM accuracy, as seen in Table \ref{ScenePrediction}.

\begin{table}[!h]
\centering 
\caption{\textbf{Ablation Study on Number of Scenes.} We show the affect that the number of scenes has on accuracy. Using 16 scenes outperforms every other option on nearly all metrics.} 

\resizebox{\columnwidth}{!}
{
\begin{tabular}{c | c || c c c c c}
\hline
\multirow{3}{*}{\bf Dataset} & \multirow{3}{*}{\bf \centering \# of Scenes} & \multicolumn{5}{c}{\bf Distance ($a_r$ [\%] @ km)} \\ 

& & \multirow{1}{1.4 cm}{\tt \bf \centering Street} & \multirow{1}{1.4 cm}{\tt \bf \centering City} & \multirow{1}{1.4 cm}{\tt \bf \centering Region} & \multirow{1}{1.4 cm}{\tt \bf \centering Country} & \multirow{1}{1.4 cm}{\tt \bf \centering Continent} \\ 

& & \bf $1$ km & \bf $25$ km & \bf $200$ km & \bf $750$ km & \bf $2500$ km \\

\hline
\hline

\multirow{3}{1.6 cm}{\tt \bf \centering Im2GPS3k\\ \cite{vo2017revisiting}} 

& 0 & 11.8 & 30.4 & 46.2 & 58.3 & 77.6 \\ 
& 3  & 12.0 & 31.7 & 47.0 & 59.8 & 78.4 \\ 

& 16 & \textbf{12.2} & \textbf{32.0} & \textbf{47.9} & \textbf{60.5} & \textbf{79.8} \\
& 365 & 11.9 & 31.8 & 47.2 & 58.5 & 78.6 \\

\hline
\hline

\multirow{3}{1.6 cm}{\tt \bf \centering YFCC26k \\ \cite{theiner2022interpretable}}  

& 0 & 8.0 & 19.8 & 30.1 & 44.6 & 62.2 \\ 
& 3  & 8.4 & 20.5 & 31.0 & 46.0 & 64.8 \\ 

& 16 & \textbf{8.7} & 21.4 & \textbf{31.6} & \textbf{47.8} & \textbf{66.2} \\
& 365 & 8.5 & \textbf{21.6} & 30.2 & 46.4 & 64.9 \\

\hline 
\hline
\end{tabular}}

\label{ScenePrediction}
\vspace{-0.2cm}
\end{table}

\subsubsection{Ablation on Number of Hierarchies}

While previous works utilized geographic hierarchies, they were either used separately \cite{vo2017revisiting} or limited to only using 3 levels of specificity \cite{pramanick2022world, muller2018geolocation}. We are the first to use more than  3 hierarchies in a combined manner. As mentioned in the main paper, our hierarchies are defined by limiting the number of training images in an S2 cell. All hierarchies have a minimum threshold of 50 images, while the maximum number of images is anywhere from 25000 to 500 depending how geographically fine-grained each class must be, specific values can be found in Table \ref{Hyperparameters}. We use at most 7 hierarchies but experiment with 1, 3, 5, and 7 on the testing datasets Im2GPS3k \cite{vo2017revisiting} and YFCC25600 \cite{theiner2022interpretable}. Each model is trained only with the number of classifiers specified by the number of hierarchies.
The ultimate goal of geo-localization is to predict the location of an image as accurately as possible. With this in mind Table \ref{Hierarchy Ablation} shows that adding more hierarchies improves geo-localization accuracy at the 1KM scale by as much as 1.5\% over the established 3 hierarchies. However, if one were not wanting to find the exact location, but instead the country or continent, then it seems that 3 hierarchies shows the best result. It appears that introducing extra fine-grained geographic classes causes our model to focus on extracting features to predict an image's location as precisely as it can at the cost of not finding features that correspond to coarser geographic hierarchies. 

\begin{table}[t!]
\centering 
\caption{\textbf{Ablation Study on Number of Hierarchies.} We show our results when varying the number of hierarchies used during training.}

\resizebox{\columnwidth}{!}
{
\begin{tabular}{c | c || c c c c c}
\hline
\multirow{3}{*}{\bf Dataset} & \multirow{3}{*}{\bf \centering \# of hierarchies} & \multicolumn{5}{c}{\bf Distance ($a_r$ [\%] @ km)} \\ 

& & \multirow{1}{1.4 cm}{\tt \bf \centering Street} & \multirow{1}{1.4 cm}{\tt \bf \centering City} & \multirow{1}{1.4 cm}{\tt \bf \centering Region} & \multirow{1}{1.4 cm}{\tt \bf \centering Country} & \multirow{1}{1.4 cm}{\tt \bf \centering Continent} \\ 

& & \bf $1$ km & \bf $25$ km & \bf $200$ km & \bf $750$ km & \bf $2500$ km \\

\hline
\hline

\multirow{3}{1.6 cm}{\tt \bf \centering Im2GPS3k\\ \cite{vo2017revisiting}} 

& 1 & 9.8 & 29.6 & 41.1 & 56.4 & 73.5 \\
& 3  & 12.8 & 34.5 & \textbf{46.1} & \textbf{61.5} & \textbf{76.7} \\
& 5 & 13.4 & 34.4 & 45.4 & 61.1 & 76.1 \\
& 7 & \textbf{14.3} & \textbf{34.8} & 45.7 & 61.3 & 76.0 \\

\hline
\hline

\multirow{3}{1.6 cm}{\tt \bf \centering YFCC26k \\ \cite{theiner2022interpretable}}  

& 1 & 6.7 & 18.2 & 29.0 & 45.2 & 64.0 \\
& 3  & 10.1 & \textbf{24.3} & 34.7 & \textbf{50.1} & \textbf{67.8} \\
& 5 & 10.2 & 24.1 & \textbf{34.8} & 50.0 & 67.7 \\
& 7 & \textbf{10.8} & 23.5 & 34.0 & 49.3 & 67.4 \\

\hline 
\hline

\multirow{4.8}{1.6 cm}{\tt \bf \centering GWS15k \\}  

& 1 & 0.0 & 0.9 & 5.7 & 21.8 & 44.0 \\
& 3  & 0.2 & 1.3 & 7.9 & \textbf{25.4} & \textbf{49.4} \\
& 5 & \textbf{0.6} & \textbf{1.7} & \textbf{8.1} & 24.3 & 48.0 \\
& 7 & 0.2 & 1.0 & 6.9 & 22.7 & 46.2 \\

\hline 
\hline
\end{tabular}}

\label{Hierarchy Ablation}
\vspace{-0.2cm}
\end{table}

\subsection{Analysis of the Google World Streets 15k dataset}

In the main paper, we discussed how previous worldwide geo-localization datasets focused heavily on tourist heavy \textit{landmarks}, ignoring systems' ability to localize more common, everyday scenes. Additionally, we showed our Google World Streets 15k is designed to capture more of these scenes, by taking random Google Street View snapshots based on a landmass-based weighting metric, rather than pulling from photography databases. 

\subsubsection{Lorenz Curves}
\begin{figure}
\centering
\includegraphics[width=1\linewidth]{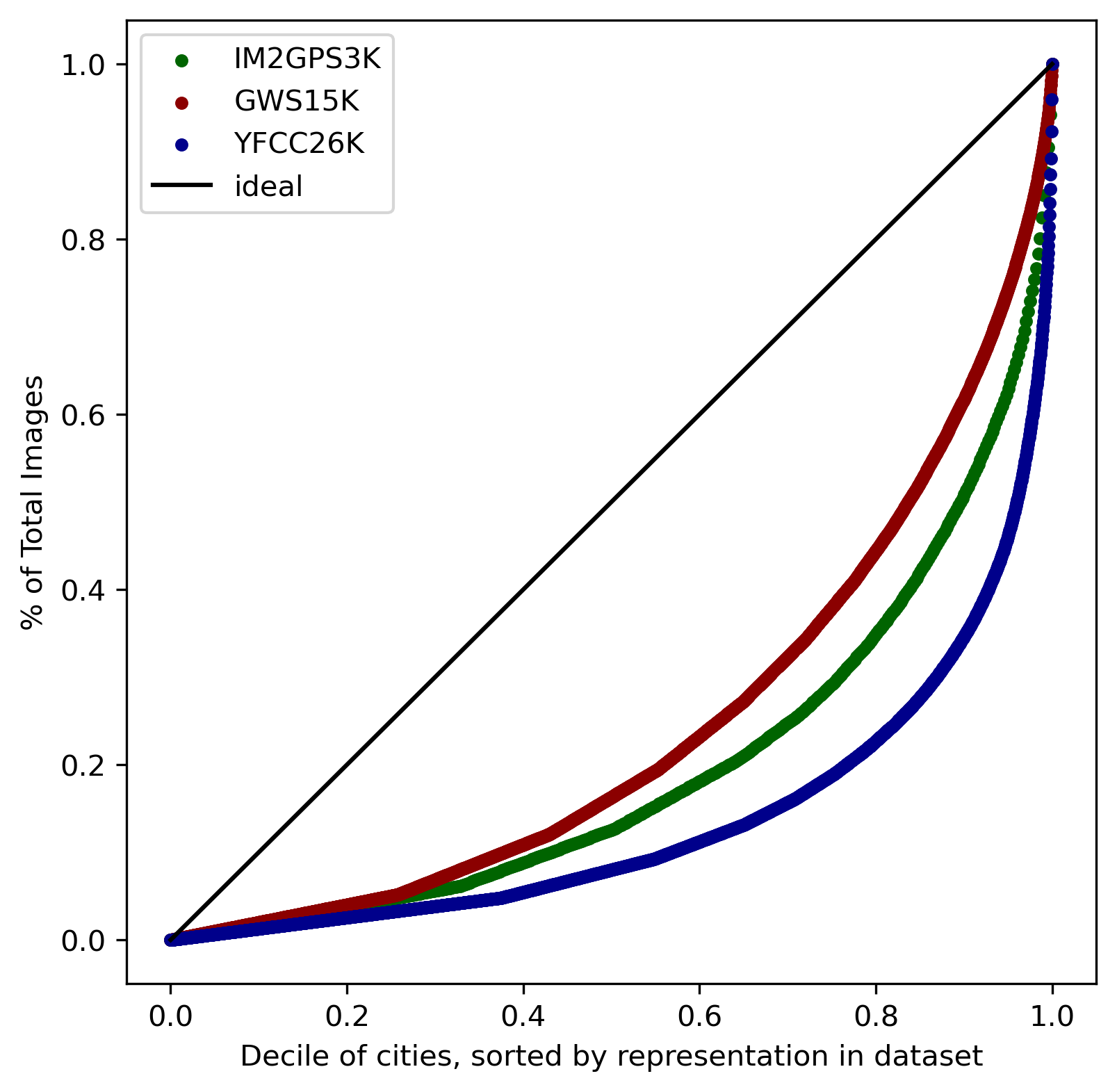}
\caption[LorenzCurves]{A comparison of the Lorenz curves of each validation dataset. The x-axis represents \textit{the botton x decile of cities, ordered by number of images} while the y-axis represents the number of images that decile contains. The black line represents perfect equality, and therefore the closer to this line, the better. While all three datasets contain some level of inequality, we see the curves of IM2GPS and YFCC rise sharply near the end, implying the most represented cities makeup a very large percentage of the dataset.}
\label{fig:Lorenz} 
\end{figure}

A Lorenz curve is a technique used to measure the distribution of some resource.\cite{lorenz1905methods} While typically used for income inequality, we can use it to demonstrate the distribution of images globally within each dataset. In our case(Fig. \ref{fig:Lorenz}), the x axis represents the cities with the bottom x\% of total images, with the y axis representing the cumulative number of images. A perfectly equal distribution would be represented by the line $y=x$. While both both datasets contain some level of inequality (as larger metropolitan cities are both larger and have more images available), the top 1\% of cities within IM2GPS3k have significantly more images than even the top 5\%.

\subsubsection{Gini Coefficient}

The Gini coefficient $G$ is a measure of inequality within some frequency distribution\cite{gates1909mutability}. It is an estimation of the area under a distribution's Lorenz curve vs the ideal case ($y=x$), thus providing a rigorous calculation of equality. Again, in our case, this distribution will be the number of images per city.  Let $i$ and $j$ represent two cities. $x_{i}$ represents the number images in city $i$. $\overline{x}$ is the average number of images per city, across all cities, serving as a normalization term. Then, we calculate $G$ as follows.
\begin{align}
G = \frac{\sum_{i=1}^{n}\sum_{j=1}^{n}\left | x_{i}-x_{j} \right |}{2n^{2}\overline{x}}
\end{align}

In the ideal case, every city has the same number of images, and therefore the difference between all cities is zero. Therefore, the resulting $G$ would also be $0$. Performing this calculation on each of our validation sets, we find IM2GPS3K's and YFCC26k's coefficients to be 0.60 and 0.73, respectively. GWS15k's coefficient is 0.51, an improvement of nearly 0.10

\subsubsection{Examples of Image Localizations}
\begin{figure*}[h]
    \centering
    \includegraphics[scale=0.33]{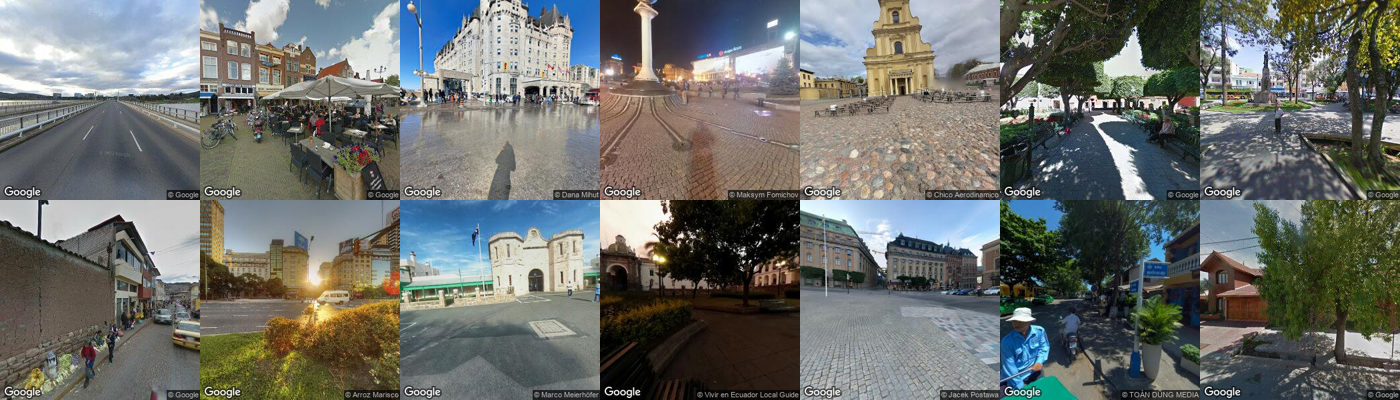}
    \caption{A random sample of GWS15k images our system correctly geo-localizes within 1KM. While well-known landmarks are greatly represented in this sample, we can also see more common locations, such as parks and a city market.}
    \label{fig:GWS15k1Km}
\end{figure*}

\begin{figure*}[h]
    \centering
    \includegraphics[scale=0.33]{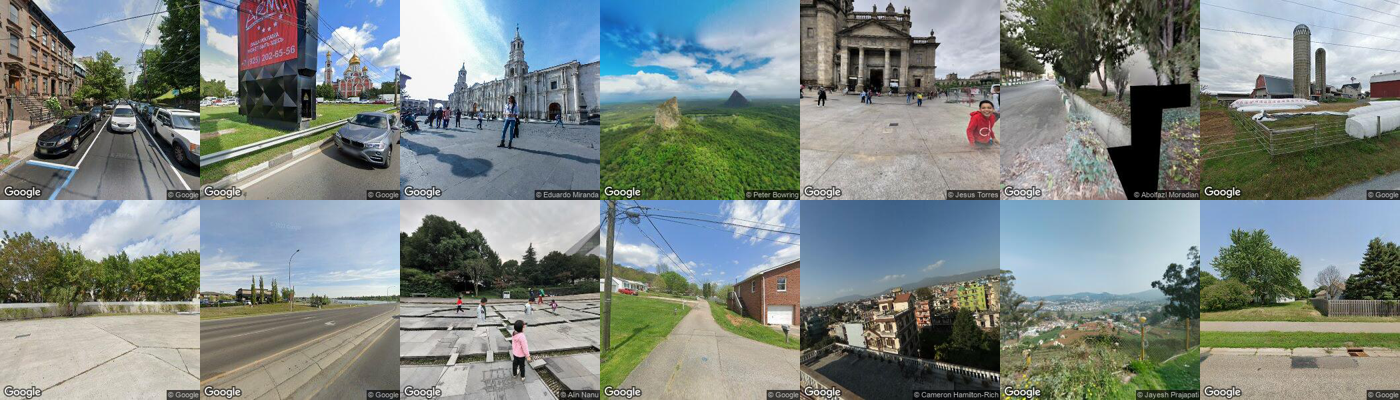}
    \caption{A random sample of GWS15k images our system correctly geo-localizes within 25KM. In this sample, we can begin to see more natural and non-urban locations. Here, our system is able to correctly identify the city of neighborhoods, as well as the rough location of rural highways with interesting structures.}
    \label{fig:GWS15k25Km}
\end{figure*}

\begin{figure*}[h]
    \centering
    \includegraphics[scale=0.33]{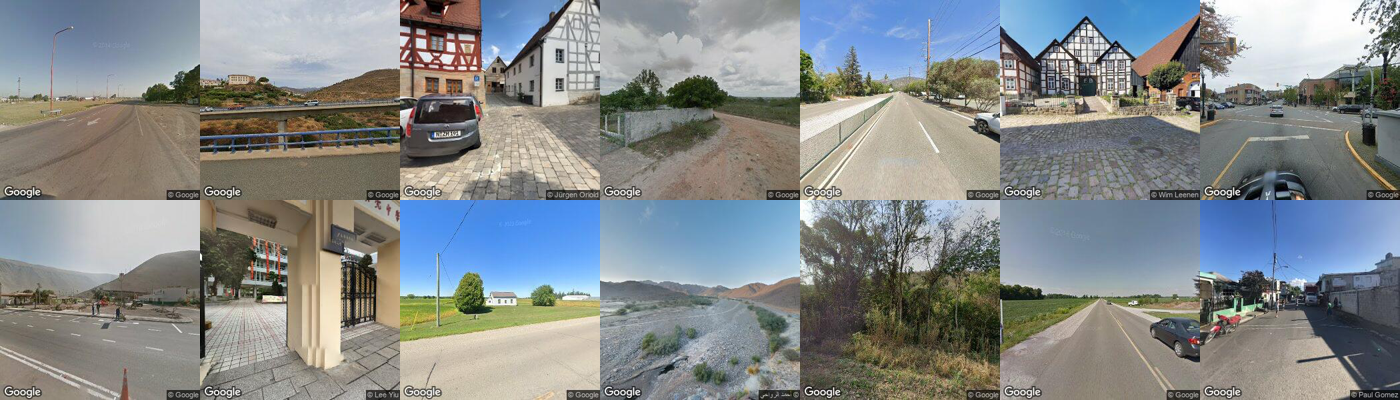}
    \caption{A random sample of GWS15k images our system correctly geo-localizes within 200KM. Here, we begin to see more challenging images. While we have two images from a city street, every other image is on an exurban road or a rural highway. Nevertheless, we can notice geography such as gravel quarries or plateaus that assists our system in identifying these locations.}
    \label{fig:GWS15k200Km}
\end{figure*}

\begin{figure*}[h]
    \centering
    \includegraphics[scale=0.33]{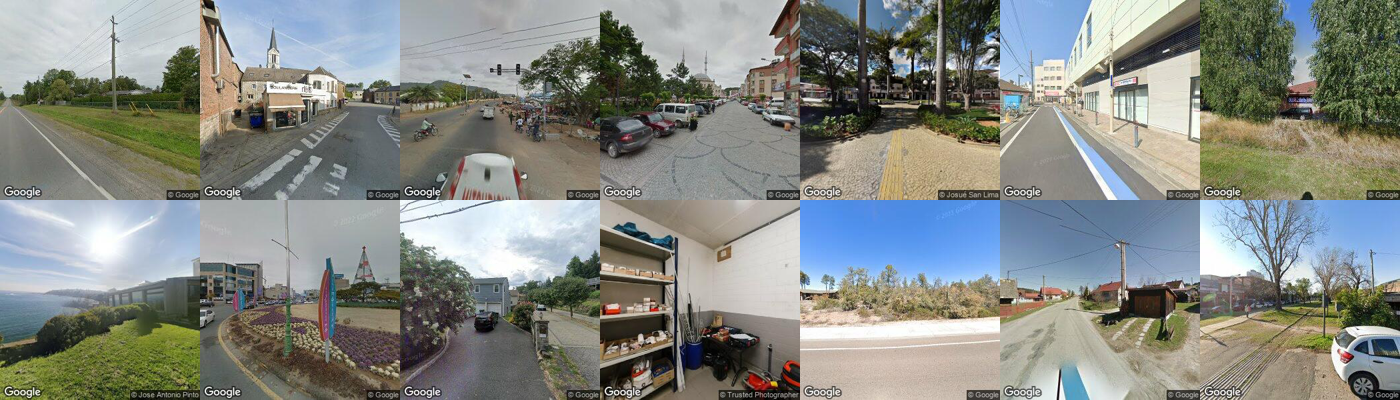}
    \caption{A random sample of GWS15k images our system correctly geo-localizes within 750KM. As this sample includes images of which the country was correctly determined, we can see examples of architecture specific to nations, but not necessarily regions. These types of structures often include types of landposts, or road signs.}
    \label{fig:GWS15k750Km}
\end{figure*}

\begin{figure*}[h]
    \centering
    \includegraphics[scale=0.33]{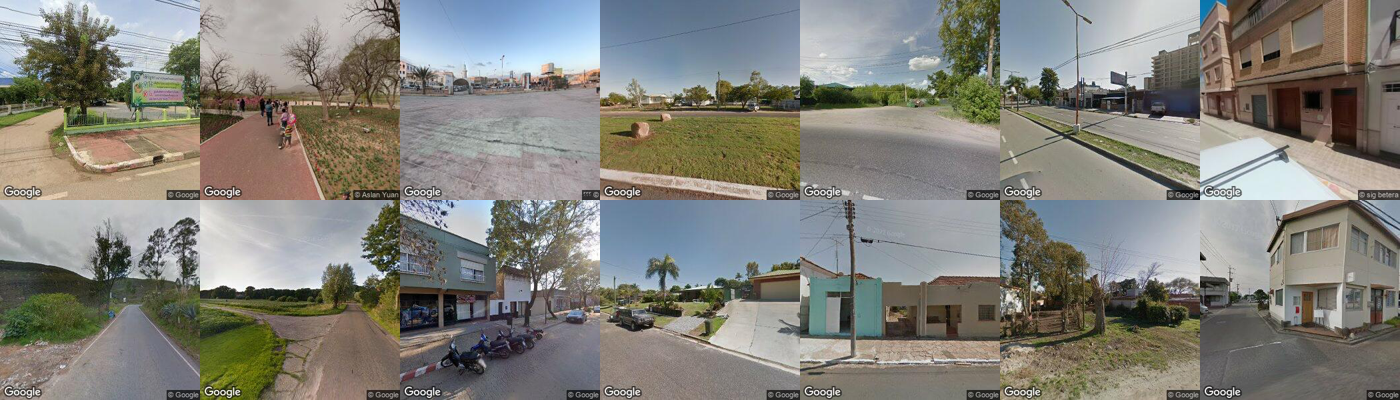}
    \caption{A random sample of GWS15k images our system correctly geo-localizes within 2500KM. As these are images of which only the continent could be determined, nearly all samples of are rural locations. These images show little differentiable geography, but enough fauna or examples of architecture to determine the continent.}
    \label{fig:GWS15k2500Km}
\end{figure*}

\begin{figure*}[h]
    \centering
    \includegraphics[scale=0.33]{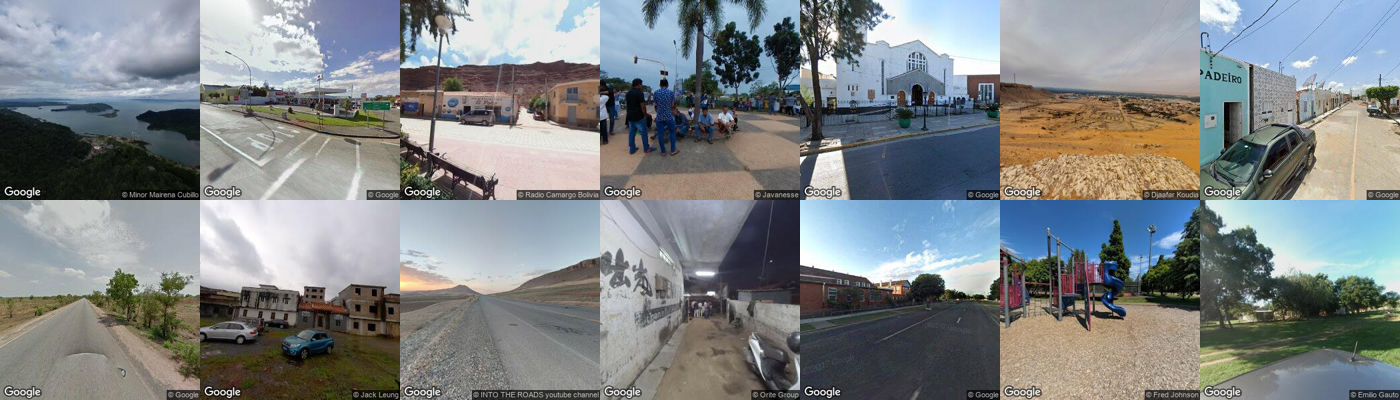}
    \caption{A random sample of failure cases, where our system's geo-localization error was over 3000KM. }
    \label{fig:GWS15kfailure}
\end{figure*}

In the following subsection, we detail a number of example localizations from our GWS15k dataset, as well as an example of failure cases. These are shown in Figures \ref{fig:GWS15k1Km}, \ref{fig:GWS15k25Km}, \ref{fig:GWS15k200Km}, \ref{fig:GWS15k750Km}, \ref{fig:GWS15k2500Km} and \ref{fig:GWS15kfailure}. As expected, strong geo-localization is commonly realized in images that contain noticeable landmarks, architecture, and geography. However, GWS15k also challenges geo-localization systems with common, everyday locations. These types of locations are also represented in locations correctly geo-localized by our model. Examining failure cases, we can see that we still struggle with locations that contain almost no man-made structures or structures not specific to a location–such as a playground.

\subsection{Review of Baselines}
\begin{itemize}
  \item \textbf{Vision Transformer (ViT)} \cite{dosovitskiy2020image} This architecture, inspired from Natural-Language Processing, breaks an image into non-overlapping squares, called patches, and feeds them through multiple layers of self-attention. For our baselines with this encoder, we use the $ViT$-$B$ architecture pre-trained on ImageNet 21k\cite{ridnik2021imagenet} and train it on MP-16 \cite{larson2017benchmarking}. For classification we use the $cls$ token output and feed it into 3 classifiers (one for each geographic hierarchy) and a classifier to predict the scene label as an ancillary task. The results of this are show in Table 5 of the main paper.

  \item \textbf{Shifted Window Transformer (Swin)} \cite{liu2021swin} Building off of ViT, Swin Tranformers re-think the self-attention step and instead perform attention only within specific windows that shift at different layers. Swin also performs a patch merging operation which lets the model learn multi-scale features. We use this architecture pretrained on ImageNet 21k \cite{ridnik2021imagenet} and train it on MP-16 \cite{larson2017benchmarking} as we did with ViT. Swin, however, does not use a $cls$ token but instead outputs a feature map of size $7\times7$. Therefore, we average pool the feature map to get one set of features, which is passed to the geographic and scene classifiers. These results are also in Table 5 of the main paper.
\end{itemize}

\subsection{Implementation Details}
\subsubsection{Hyperparameters}
Our model is trained for 40 epochs with a batch-size of 512. We utilize Stochastic Gradient Descent with an initial learning rate of 0.01, momentum of 0.9, and weight decay of 0.0001. Our encoder is pretrained on ImageNet\cite{ridnik2021imagenet} and we use Cross-Entropy for all of our losses. We outline the training parameters for our model in Table \ref{Hyperparameters}.
\begin{table}[t!]
\centering 
\caption{\textbf{Training parameters for our model}}

\resizebox{\columnwidth}{!}
{
\begin{tabular}{|c || c|}
\hline
\multirow{1}{*}{\bf Hyperparameter} & \multirow{1}{*}{\bf \centering Value} \\

\hline

Batch-size & 512 \\
Epochs  & 40 \\
Optimizer & SGD \\
Learning Rate & 0.01 \\
Momentum & 0.9 \\
Weight Decay & 0.0001 \\
Hierarchy Losses & Cross-Entropy \\
Scene Loss & Cross-Entropy \\
Scheduler & MultiStepLR \\
Milestones & [4, 8, 12, 13, 14, 15]\\
Gamma & 0.5 \\
Maximum Images per Class & [25000, 10000, 5000, 2000, 1000, 750, 500] \\
Minimum Images per Class & 50 \\
Classes per Hierarchy & [684, 1744, 3298, 7202, 12893, 16150, 21673] \\

\hline

\end{tabular}}

\label{Hyperparameters}
\vspace{0.2cm}
\end{table}
\\
\subsubsection{Augmentations}
We utilize the following augmentations during training:
\begin{itemize}
    \item Random Affine (1-15 degrees)
    \item Color Jitter (brightness=0.4, contrast=0.4, saturation=0.4, hue=0.1)
    \item Random Horizontal Flip (probability=0.5)
    \item Resize (256$\times$256)
    \item RandomCrop (224$\times$224)
    \item Normalization
\end{itemize}

During Evaluation we use:
\begin{itemize}
    \item Resize (256$\times$256)
    \item TenCrop (224$\times$224)
    \item Normalization
\end{itemize}

TenCrop is an augmentation technique that, given an image, returns the center and corner crops as well as the horizontally flipped version of each of those crops.

\subsection{Qualitative Analysis}
\begin{figure*}[h]
    \centering
    \includegraphics[scale=0.2]{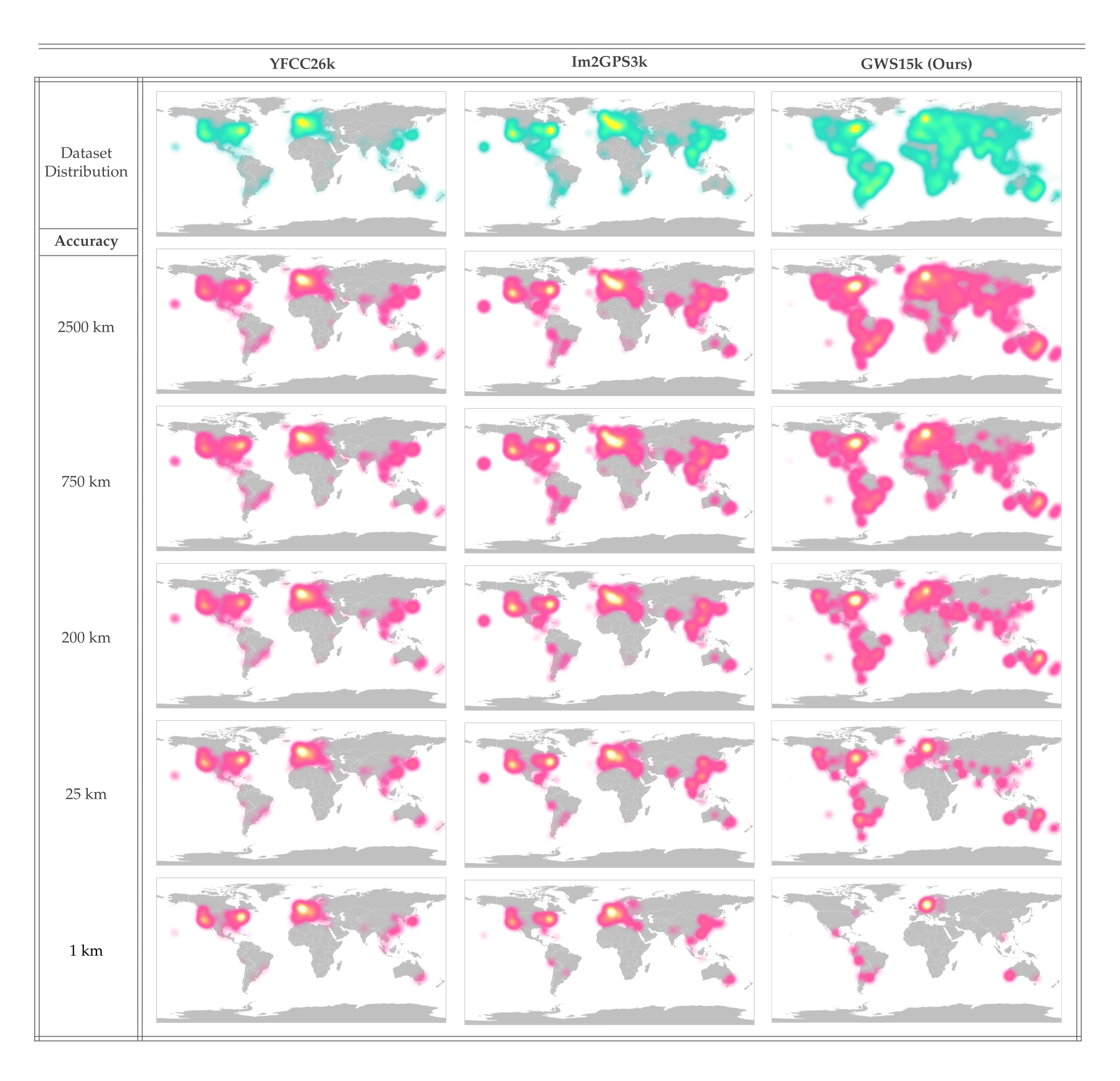}
    \caption{A visualization of the distribution of our model's correct predictions for different accuracy thresholds and testing datasets. By observing the predictions of our model on GWS15k we can note that we are capable of identifying places that are underrepresented in YFCC26k and Im2GPS3k. It is important to note that although our model's performance is being limited by the training set–whose distribution is comparable to that of YFCC26k–we accurately geo-localize images in areas that are not densely covered with high precision (e.g. Western Asia, South America, and Central Australia).}
    \label{fig:MutliAccMaps}
\end{figure*}

In figure \ref{fig:MutliAccMaps} we show the locations that our model predicts within each distance threshold for Im2GPS3k, YFCC26k, and GWS15k. The overall dataset distributions are shown for reference. We observe that our method can accurately predict images in locations that other datasets leave out. This shows our model's capabilities better by ensuring we test on images all around the Earth and aren't biased towards North America and Europe.

\begin{figure*}[h]
    \centering
    \includegraphics[width=17cm, scale=0.7]{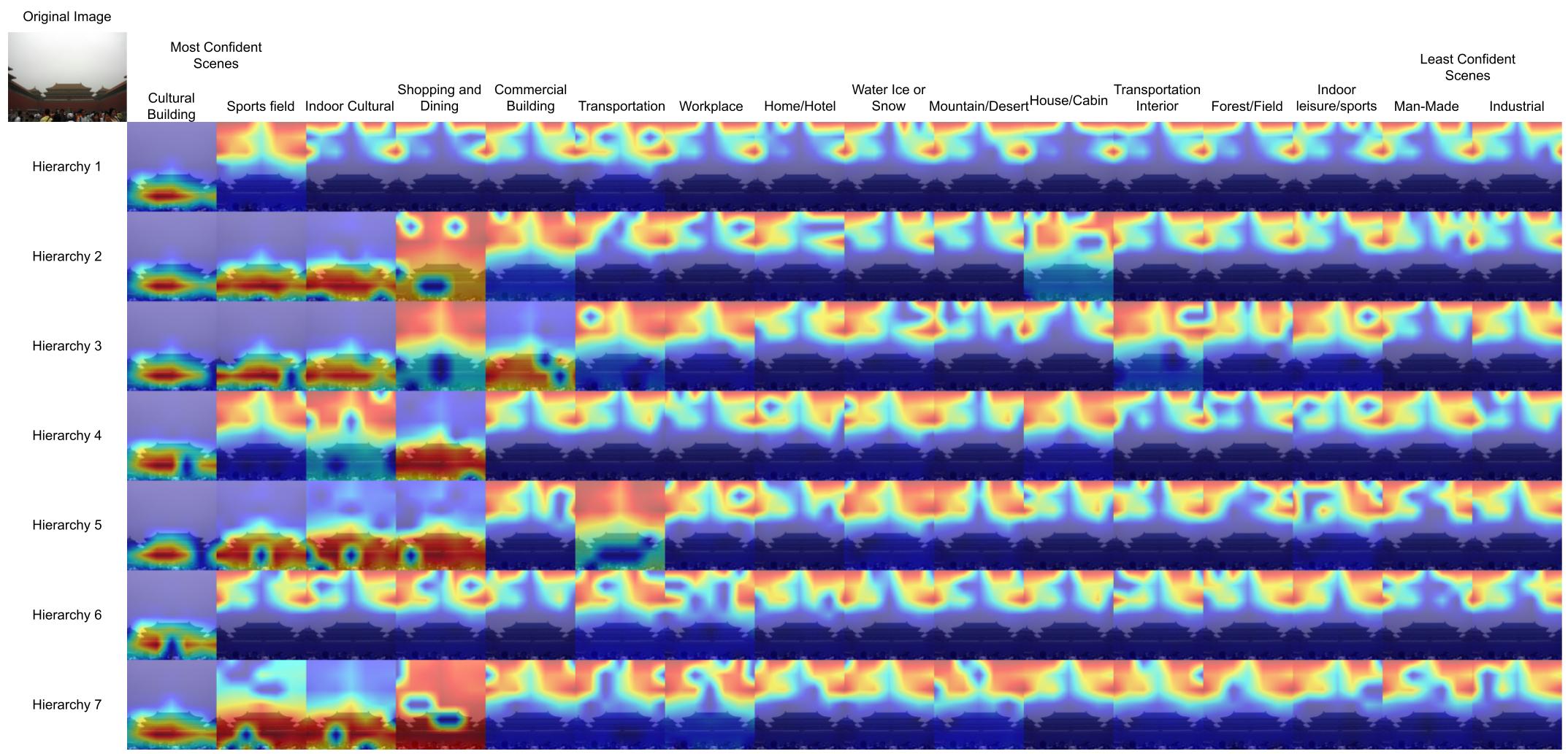}
    \caption{A visualization of all the attention maps for an image of The Palace Museum in Beijing from Im2GPS3k that we predict within 0.32 KM. We see that the left most column, which represents the query we use for classification, has a far more direct attention map then the incorrect scene queries.}
    \label{fig:im2gps3k1Km}
\end{figure*}

\begin{figure*}[h]
    \centering
    \includegraphics[width=17cm]{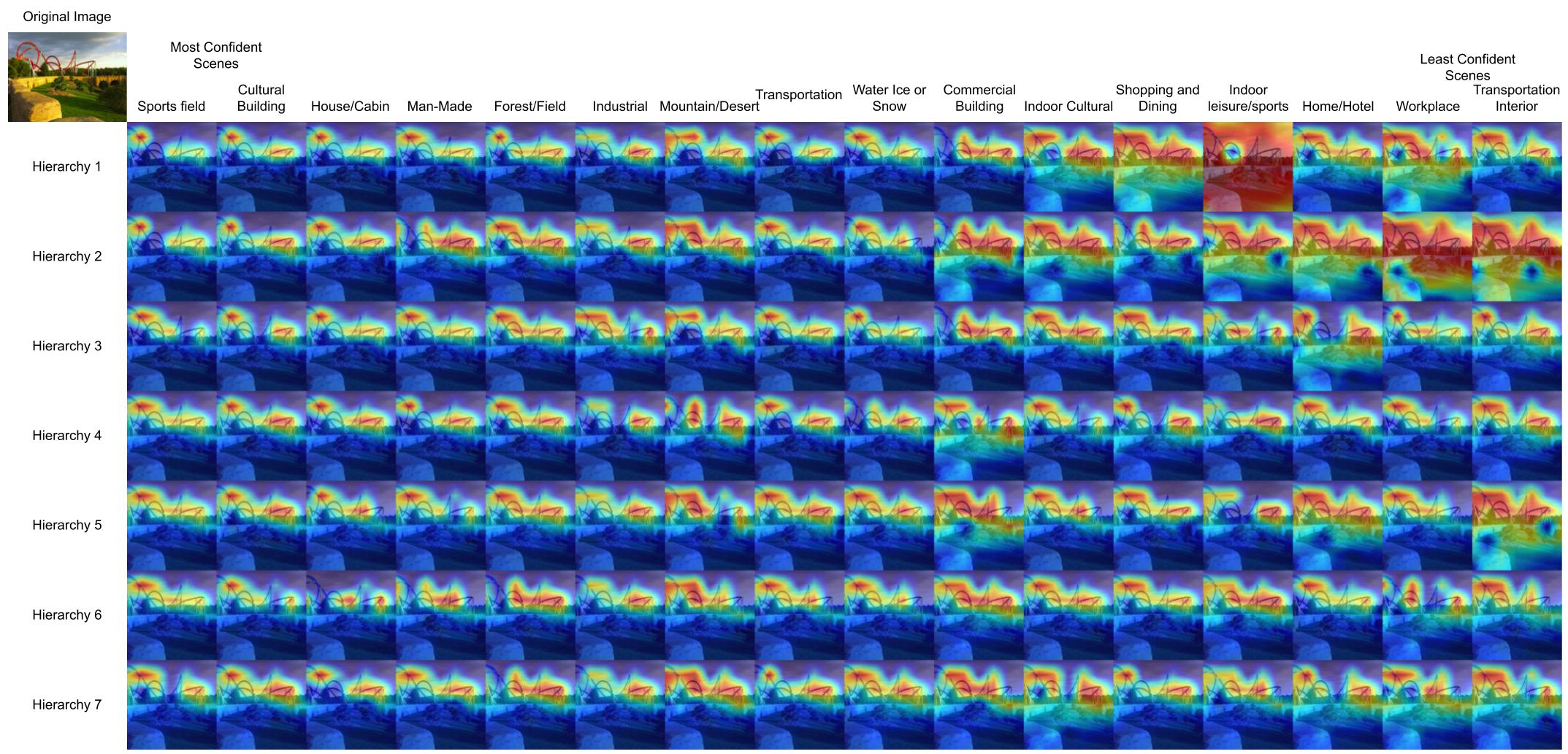}
    \caption{A visualization of all the attention maps for a failure-case image from Barcelona, Spain in Im2GPS3k that we mispredict by 5284 KM. We see that nearly all of the queries are focusing on the roller coasters seen in the background. Our model was not able to find features in the image specific enough to a scene or hierarchy in order to geolocalize it.}
    \label{fig:im2gps3k2501Km}
\end{figure*}

\begin{figure*}[h]
    \centering
    \includegraphics[width=17cm]{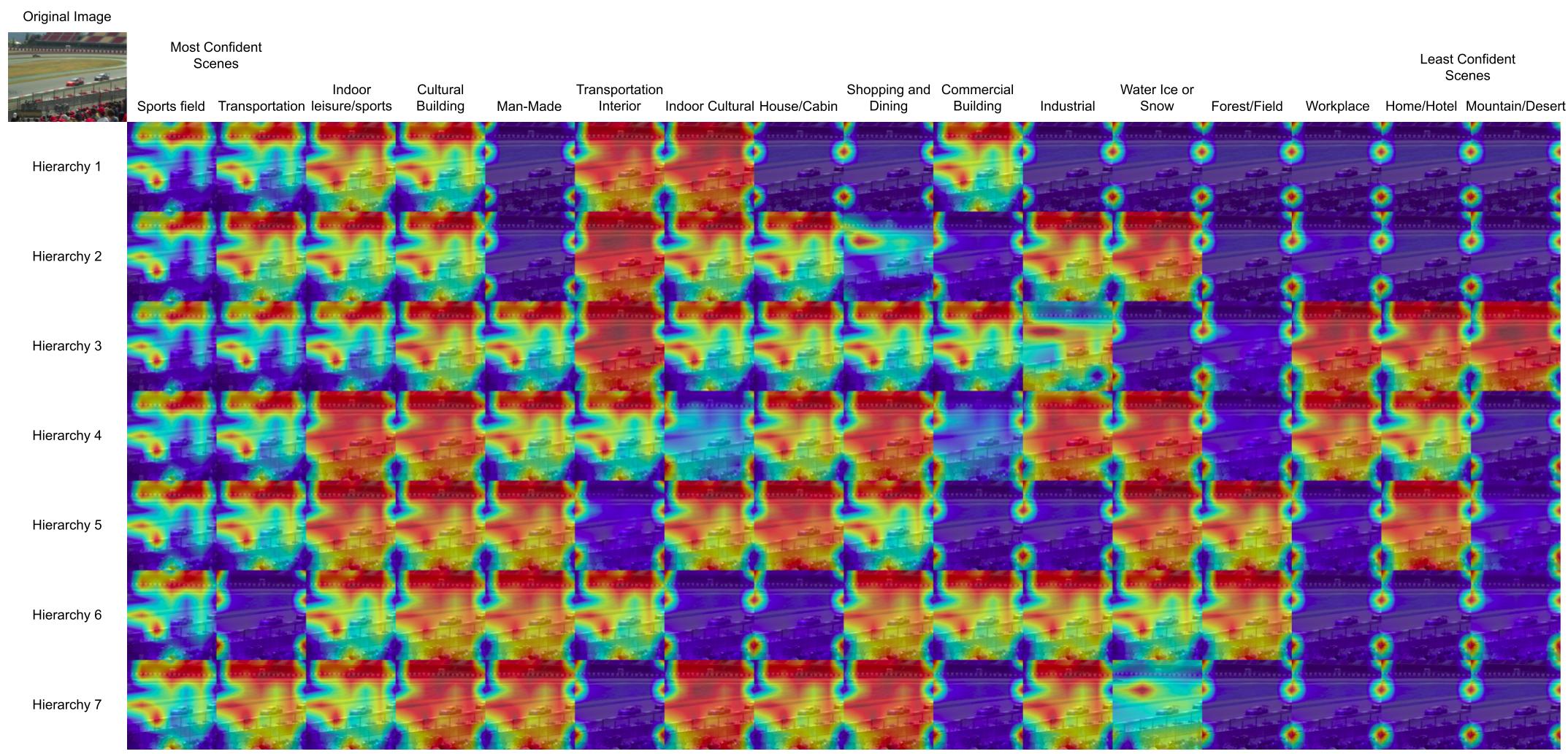}
    \caption{A visualization of all the attention maps for a success-case image at Catalunya Circuit in Barcelona from YFCC26k that we predict within 0.75 KM. Note that the queries used for classification focus on the stands in the background and part of the race track, while the less confident scenes focus either on the corners or generally about the entire image.}
    \label{fig:yfcc26k1Km}
\end{figure*}

\begin{figure*}[h]
    \centering
    \includegraphics[width=17cm]{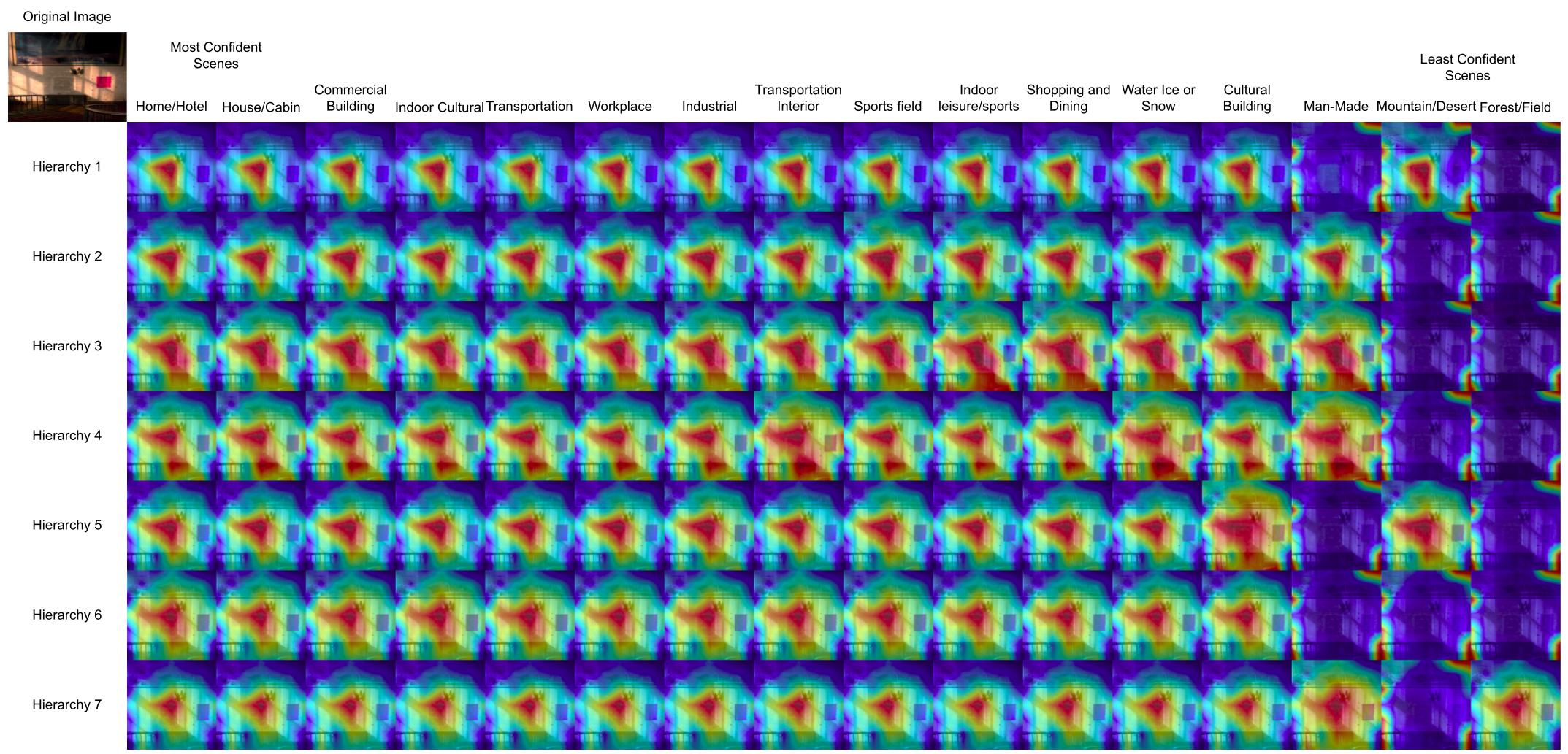}
    \caption{A visualization of all the attention maps for a failure-case image from Anshan, China in YFCC26k that we mispredict by 8442 KM. We see here that almost all scenes show identical attention maps no matter how confident we are in that scene's prediction. Note that this image is also an indoor image of a wall inside this building so we expect this to be an especially difficult image to localize unless images of the same wall exist in the training set.}
    \label{fig:yfcc26k2501Km}
\end{figure*}

\begin{figure*}[h]
    \centering
    \includegraphics[width=17cm]{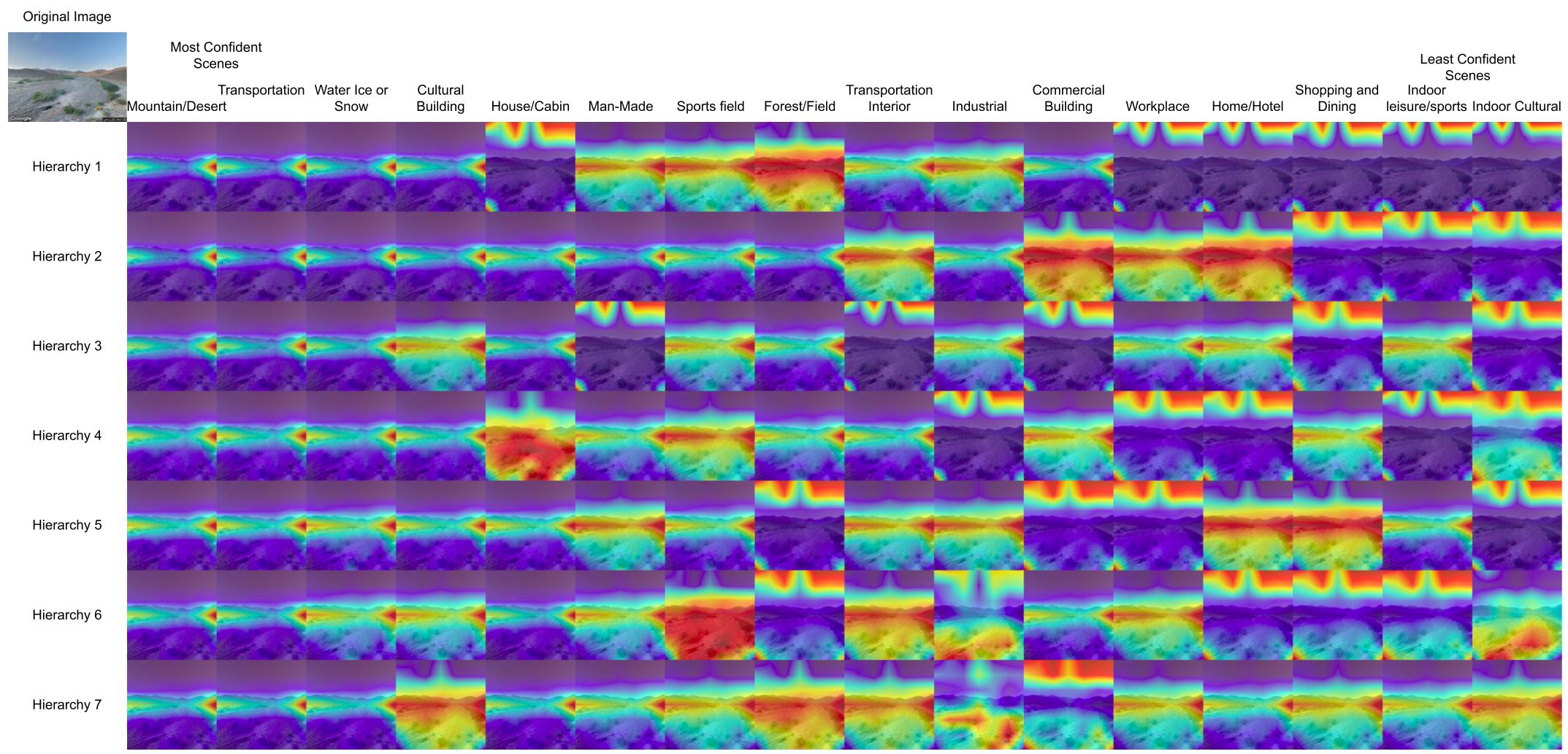}
    \caption{A visualization of all the attention maps for a success-case image from Samail, Oman in GWS15k that we predict within 7.3 KM. We see that the most confident scene queries are consistently focusing on the mountains in the background while the less confident queries do not, showing that our scene selection process helps our model get the best features for geo-localization.}
    \label{fig:GWS15k1Km-att}
\end{figure*}

\begin{figure*}[h]
    \centering
    \includegraphics[width=17cm]{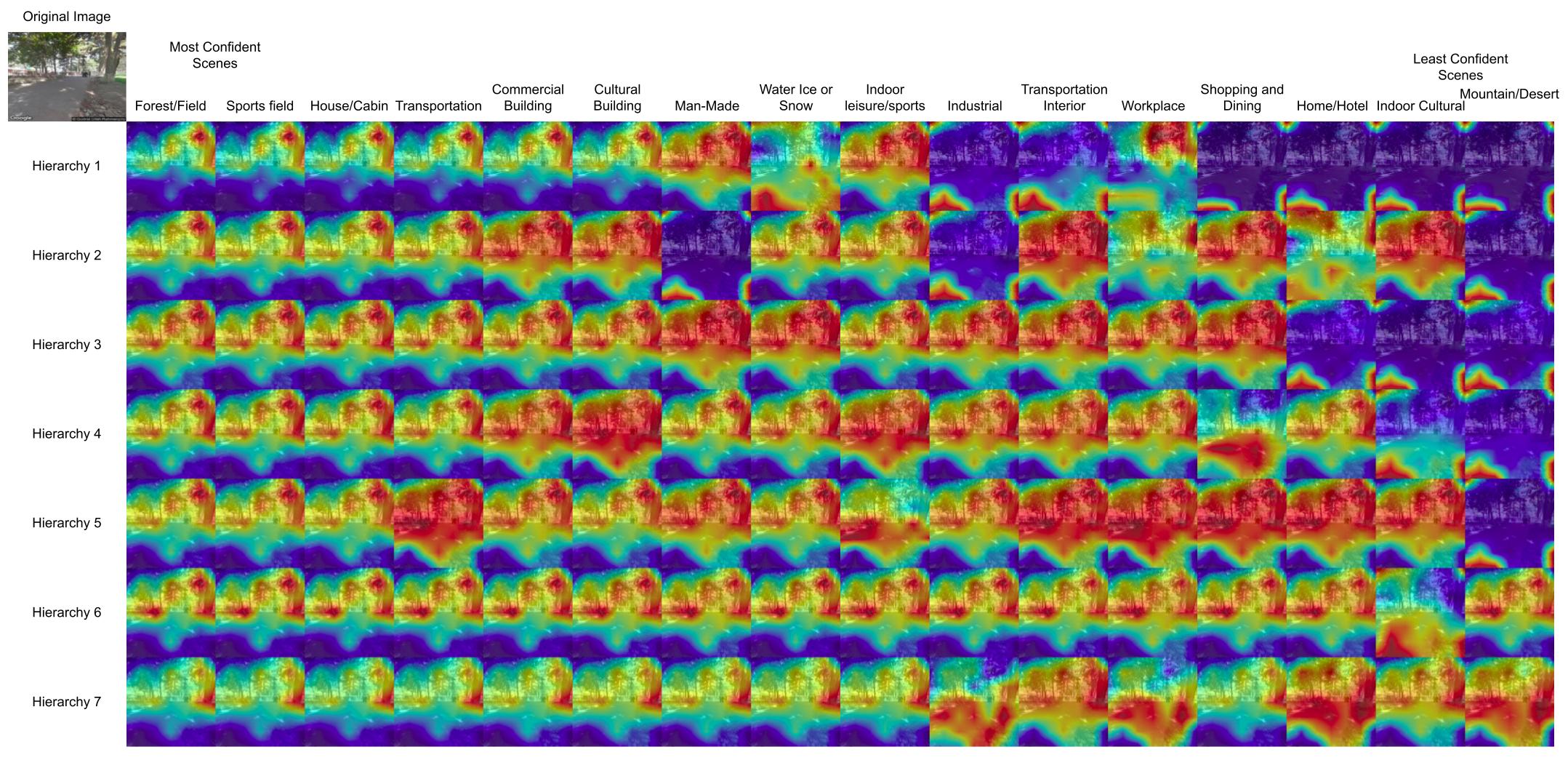}
    \caption{A visualization of all the attention maps for a failure-case image from Mashal University, Afghanistan in GWS 15k that we mispredict by 3490 KM. We see that in this case the queries we would use (the leftmost column) share similar attention maps to the less confident scenes, meaning we could not distinguish this image's features well enough.}
    \label{fig:GWS15k2501Km}
\end{figure*}

In figures \ref{fig:im2gps3k1Km}, \ref{fig:im2gps3k2501Km}, \ref{fig:yfcc26k1Km}, \ref{fig:yfcc26k2501Km}, \ref{fig:GWS15k1Km-att}, and \ref{fig:GWS15k2501Km} we provide attention maps from Im2GPS3k \cite{vo2017revisiting}, YFCC26k \cite{theiner2022interpretable}, and GWS15k. We provide one success and one failure case for each of these datasets. The attention maps are created by looking at the first attention head in the final layer of cross-attention in our Hierarchy Dependent Decoder. This shows the relation between the decoder queries and the image patches from our encoder. The attention maps for every hierarchy and scene query are shown as well as the original image in the top left for reference.



\clearpage

\end{document}